\pdfoutput=1

\documentclass[11pt]{article}

\usepackage[]{acl}

\usepackage{times}
\usepackage{latexsym}

\usepackage[T1]{fontenc}

\usepackage[utf8]{inputenc}

\usepackage{microtype}

\usepackage{inconsolata}

\usepackage{inconsolata}
\usepackage{graphicx}
\usepackage{colortbl}
\usepackage{bbm}
\usepackage{amsfonts}
\usepackage[normalem]{ulem}
\useunder{\uline}{\ul}{}
\usepackage{amssymb}
\usepackage{amsfonts}
\usepackage{framed}
\usepackage{color}
\usepackage{multicol}
\usepackage{multirow}
\usepackage{makecell}
\usepackage{amsmath}
\usepackage{booktabs}
\usepackage{array}
\usepackage{paralist}
\usepackage{algorithm}
\usepackage{bbm}
\usepackage{dsfont}
\usepackage{adjustbox}
\usepackage{diagbox}
\usepackage{algpseudocode}
\usepackage{multicol}
\usepackage{wrapfig}
\usepackage{enumitem}
\usepackage{caption}
\usepackage[most]{tcolorbox}
\usepackage{svg}
\usepackage{arydshln}
\usepackage{arydshln}
\usepackage{bm}

\title{Preference-Guided Reflective Sampling for Aligning Language Models}

\author{Hai Ye \\ 
Department of Computer Science \\ 
National University of Singapore \\
\texttt{yehai}\texttt{@comp.nus.edu.sg} \\\And
Hwee Tou Ng \\ 
Department of Computer Science \\ 
National University of Singapore \\
\texttt{nght}\texttt{@comp.nus.edu.sg}
}

\begin{document}
\maketitle

\begin{abstract}

Iterative data generation and model re-training can effectively align large language models~(LLMs) to human preferences. The process of data sampling is crucial, as it significantly influences the success of policy improvement. Repeated random sampling is a widely used method that independently queries the model multiple times to generate outputs. In this work, we propose a more effective sampling method, named \emph{Preference-Guided Reflective Sampling (PRS)}. Unlike random sampling, \emph{PRS} employs a tree-based generation framework to enable more efficient sampling. It leverages adaptive self-refinement techniques to better explore the sampling space. By specifying user preferences in natural language, \emph{PRS} can further optimize response generation according to these preferences. As a result, \emph{PRS} can align models to diverse user preferences. Our experiments demonstrate that \emph{PRS} generates higher-quality responses with significantly higher rewards. On AlpacaEval and Arena-Hard, \emph{PRS} substantially outperforms repeated random sampling in best-of-$N$ sampling. Moreover, \emph{PRS} shows strong performance when applied in iterative offline RL training\footnote{Source code of this paper is available at \url{https://github.com/nusnlp/PRS}.}.

\end{abstract}

\section{Introduction}
Large language models (LLMs) have made significant advances~\cite{radford2019language,brown2020language,ChatGPT}. These models are typically aligned with human expectations through fine-tuning. This is achieved by using reinforcement learning from human feedback (RLHF), which mitigates the generation of harmful, biased, or irrelevant outputs~\cite{perez2022red}. Both online and offline RL methods have been explored for RLHF~\cite{schulman2017proximal,gulcehre2023reinforced,rafailov2024direct}. Iterative offline training provides a more efficient alternative than online training, by allowing outputs to be pre-generated and reused to facilitate iterative improvements in policy.

\begin{figure}[t]
\centering
\includegraphics[width=\columnwidth]{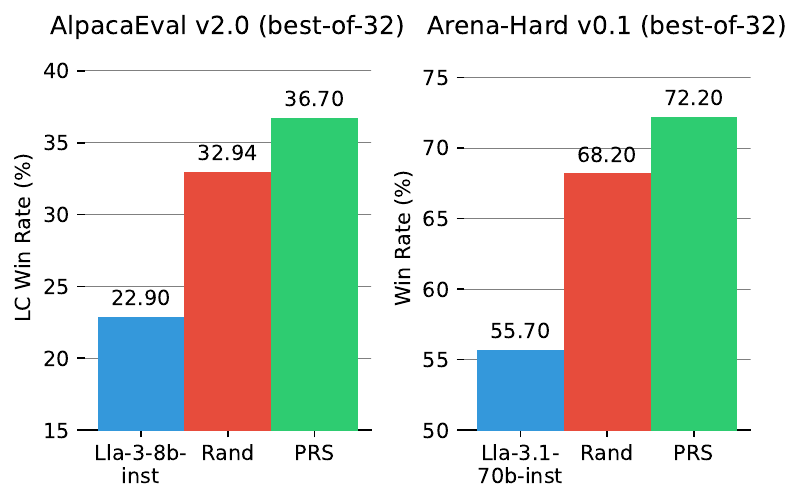} 
\caption{Performance comparison of \emph{PRS}~(ours) and repeated random sampling~(Rand) on AlpacaEval v2.0 and Arena-Hard v0.1 using best-of-32 sampling. Each prompt samples 32 responses using Rand or \emph{PRS} and the response with the highest reward is kept for evaluation.
}
\label{fig:alpaca_eval_prs_rand}
\end{figure}

\begin{figure*}[t]
\centering
\includegraphics[width=\linewidth]{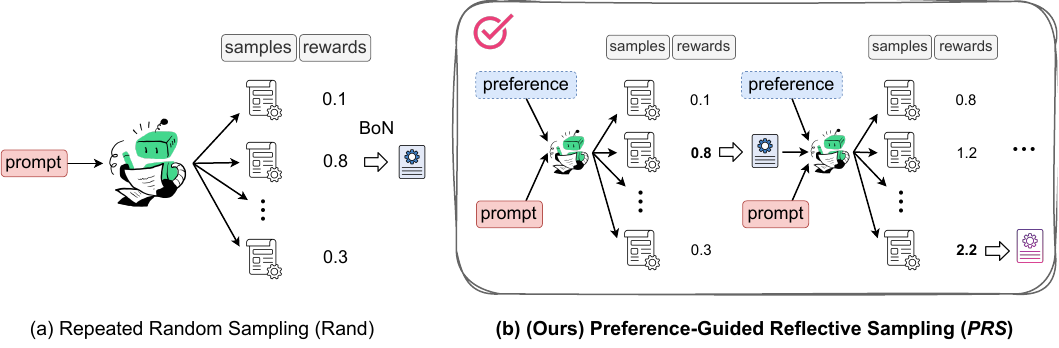} 
\caption{Comparison of repeated random sampling and our method \emph{PRS}. \emph{PRS} adopts a tree-based generation framework that learns to adapt and adjust its outputs by reflecting on its already generated data. It can incorporate a specific user preference to optimize responses that align with it. Adjusting preferences will generate tailored responses. For random sampling, it generates samples independently and can use the best-of-$N$~(BoN) method to find the best sample. Both methods share the same sampling budget, which samples the same number of responses for each prompt.
}
\label{fig:compare_rand_prs}
\end{figure*}

Effective data sampling is crucial for iterative model re-training, as it directly influences the effectiveness of the resulting policy~\cite{gulcehre2023reinforced}. 
Repeated random sampling~(as shown in Fig.~\ref{fig:compare_rand_prs}) is an effective method and has been widely used for data generation in previous work~\cite{gulcehre2023reinforced}. It independently calls the model multiple times to get samples. Then higher-quality data will be maintained to update the policy model. 
However, the vast output space compromises its efficiency since the inherent randomness may result in inefficient exploration in the sampling space. Also, the simple generation strategy cannot learn from and adapt dynamically based on previously generated samples. Furthermore, with only the supervision of the reward model, it is hard to optimize the outputs to align to diverse and personalized preferences. 

We propose a new sampling method named \emph{Preference-Guided Reflective Sampling (PRS)} to improve data generation. Different from random sampling, \emph{PRS} employs a tree-based generation framework to balance exploration and exploitation throughout the generation process~(see Fig.~\ref{fig:compare_rand_prs}). It learns to adapt and adjust its outputs by reflecting on its already generated data so that it can improve the sampling of future samples. Furthermore, by using a preference described in natural language, \emph{PRS} can optimize the response toward this explicit preference. The user preference is incorporated as an additional sampling context, guiding the model toward more relevant directions and minimizing unnecessary exploration. As a result, it achieves more efficient sampling and can also generate samples aligned to diverse preferences.

We study preference-controlled text generation for the task of instruction following and keyword-focused document summarization. In our experiments, we first evaluate \emph{PRS} against various baselines in generating training samples with diverse policy models~($\S$~\ref{sec:compare_sampling}). In $\S$~\ref{sec:offline_RL}, we investigate its application for aligning LLMs to adhere to explicit preferences provided in the inputs using offline RL training. 
We further explore preference adaptation, toxicity reduction, and other areas in $\S$~\ref{sec:further-analysis}.  
Our contributions in this work are as follows:
\begin{itemize}[itemsep=-5pt, topsep=0pt, leftmargin=0.3cm]
    \item We introduce \emph{PRS}, a novel sampling method to improve data generation. \emph{PRS} is capable of generation tailored to different preferences. 
    
    \item Experiments with 9 policy models show that \emph{PRS} generates training data with higher rewards. On AlpacaEval and Arena-Hard, \emph{PRS} achieves better performance than repeated random sampling in the best-of-$N$ setting~(Fig.~\ref{fig:alpaca_eval_prs_rand}).

    \item With extensive offline RL training, the outcomes across multiple benchmarks, e.g., AlpacaEval~\cite{alpaca_eval} highlight the effectiveness of \emph{PRS}.
    
    \item Further analysis demonstrates \emph{PRS}'s superior performance in preference adaptation. 
\end{itemize}

\begin{figure*}[t]
   \centering
\includegraphics[width=0.9\linewidth]{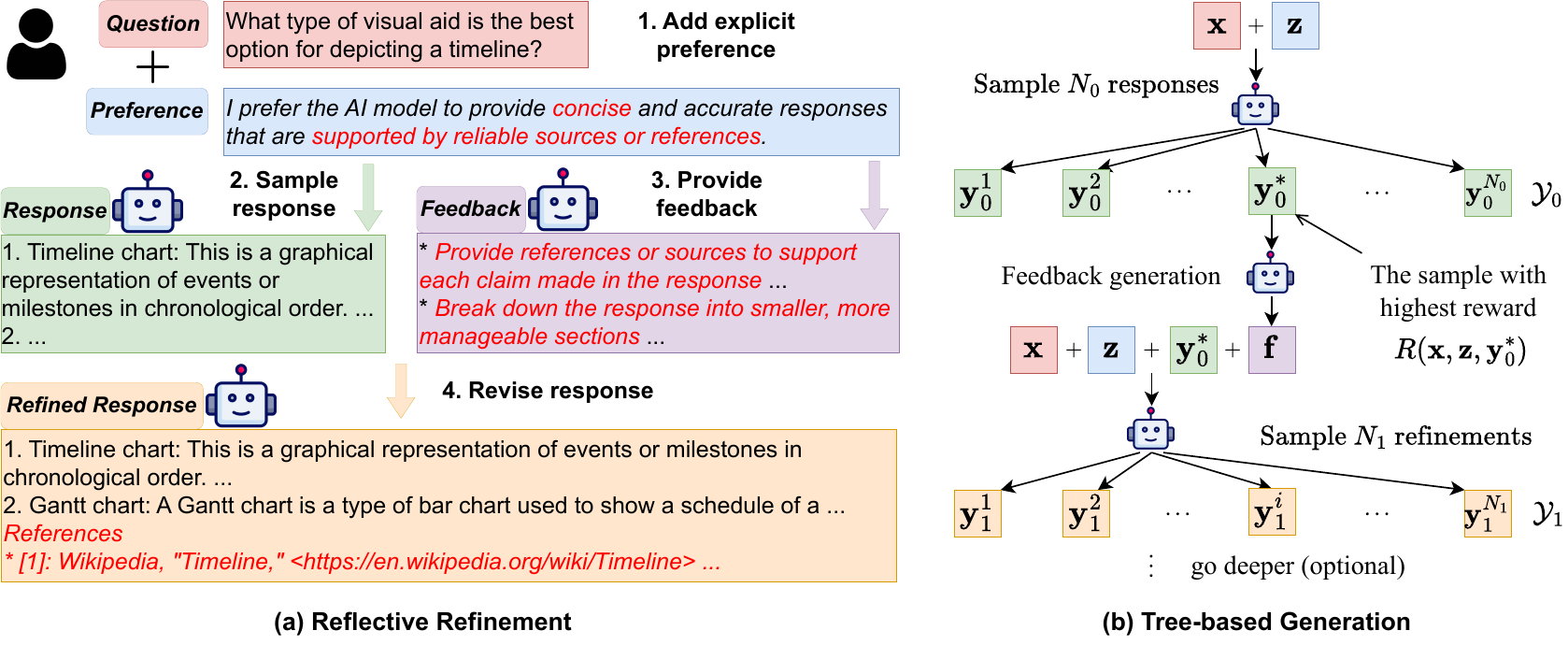} 
\vspace{-5pt}
\caption{\textbf{\emph{PRS}}: (a) Example: A user requests a brief response with supporting references. The initial response lacks references. After feedback, the revised response includes appropriate references. (b) A preference \(\bm{z}\) is added to the input \(\bm{x}\). The process begins by sampling \(N_0\) initial responses \(\mathcal{Y}_0\), from which the optimal response \(\bm{y}^*_0\) is selected using a reward model \(R\). Then feedback $\bm{f}$ is generated, leading to the sampling of \(N_1\) refinements \(\mathcal{Y}_1\) to enhance \(\bm{y}^*_0\). Finally, \(\mathcal{Y}_0\) and \(\mathcal{Y}_1\) are merged. Optionally, new refinements may be sampled based on the current best response.}
\label{fig:pres}
\end{figure*}

\section{Related Work}
Offline RL offers an efficient alternative to online RL \cite{schulman2017proximal}. 
\citet{dong2023raft}, \citet{gulcehre2023reinforced}, and \citet{rafailov2024direct} emphasize data generation and model refinement.  
Repeated random sampling is a simple but effective method for data generation. \citet{brown2024large} demonstrate that scaling inference compute can significantly improve the model performance in problem solving. 
\citet{DBLP:journals/corr/abs-2212-08073} leverage the LLM's reflection capacity to continuously refine model responses. However, they only focus on harmless responses, whereas our work is applicable across a broad spectrum of preferences. Moreover, different from ours, their work does not aim to improve data sampling for RL training. \citet{feng2023alphazero} use Monte Carlo tree search (MCTS) with token-level rewards, but ours employs sequence-level rewards based on cost-effective tree-based generation, with input preferences to guide the generation. \citet{scheurer2023training} advocate for training models using human language feedback, but we employ the model itself to generate language feedback. A more detailed discussion of the related work is in Appendix~\ref{app:related_work}.

\section{Preliminaries}\label{sec:preliminary_offline_RL}

\noindent{\textbf{Offline RL.}} \ RLHF utilizes human feedback to fine-tune a pre-trained LLM with human preferences. The preference human feedback can be utilized to train a reward model $R(\bm{x}, \bm{y})$, given an input $\bm{x}$ and a response $\bm{y}$. Following \citet{gulcehre2023reinforced}, we employ offline RL to synchronize the LLM policy with the trained reward model. This process, beginning with the policy initialized by supervised fine-tuning (SFT) on labeled data, involves iterative cycles of data generation and model re-training. 

The policy of the LLM, $\pi_{\theta}$, parameterized by $\theta$, produces a response $\bm{y}$ given the input $\bm{x}$, i.e., $\bm{y} \sim \pi_{\theta}(\bm{y}|\bm{x})$. Using the labeled data $\mathcal{D}_0$, the LLM is trained with the negative log-likelihood (NLL):
\begin{equation}
    \mathcal{L}_{NLL} = \mathbb{E}_{(\bm{x}, \bm{y}) \sim \mathcal{D}_0} \big[- \log \pi_{\theta}(\bm{y}|\bm{x}) \big]
    \label{eq:nll}
\end{equation} 
Then it repeats data generation and model re-training to align the language model:
\begin{enumerate}[itemsep=-5pt, topsep=0pt, leftmargin=0.5cm]
    \item \textbf{Data Generation:} Each iteration leverages the policy from the previous cycle to generate $N$ responses $\bm{y}$ for each input $\bm{x}$ in the unseen dataset $\mathcal{U}_k$. The reward model, $R(\bm{x}, \bm{y})$, evaluates these responses to generate rewards. Best-of-$N$ strategy or a reward threshold is used to identify the high-quality examples.
    \item \textbf{Model Re-training:} The newly generated data, along with all prior data, is used to refine the model in the subsequent re-training phase.
\end{enumerate}

\section{Method}
We aim to improve the data generation process to enhance offline RL training. We first introduce \emph{Preference-Guided Reflective Sampling (PRS)}, and then study the task of preference-controlled instruction following using offline RL training.

\subsection{Preference-Guided Reflective Sampling}
\emph{PRS} aims to optimize the response aligned to a given user preference described in natural language. The user preference describes the desired model output, such as \emph{conciseness}. Let $\bm{z}$ denote a specific preference, exemplified by statements like ``\emph{I prefer the response to be concise.}'' or ``\emph{Can you give me a response without wordy explanations?}''. \emph{PRS} aims to generate the responses aligned to the preference $\bm{z}$.

Initially, we sample a response $\bm{y}_0$ conditioned on both $\bm{x}$ and $\bm{z}$, by appending $\bm{z}$ to the input $\bm{x}$. Subsequently, we engage the LLM policy in a process of self-reflection, aiming to iteratively refine $\bm{y}_0$ to better align with the stated preference. Given the independence of preference $\bm{z}$ and input $\bm{x}$, we redefine the generation process of $p(\bm{y}|\bm{x})$ as:
\begin{equation}
\begin{aligned}
    p(\bm{y}|\bm{x})  = \sum_{\bm{z}, \bm{y}_0, \bm{f}} p(\bm{z}) \times \underbrace{\pi_{\theta}(\bm{y}_0|\bm{x}, \bm{z})}_\text{Initial Sampling} \times  \ \ \ \ \  \ \ \ \ \  \\   \underbrace{ \pi_\theta(\bm{f} | \bm{x}, \bm{z}, \bm{y}_0) \times \pi_\theta(\bm{y}| \bm{x}, \bm{z}, \bm{y}_0, \bm{f})}_\text{Reflective Refinement} 
    \label{eq:pres}
\end{aligned}
\end{equation}
where for the reflective refinement, the model first generates language feedback $\bm{f}$ for the output $\bm{y}_0$, then revises the output by incorporating the feedback to obtain a new response $\bm{y}$ (see Fig.~\ref{fig:pres}a). Promoting the model to provide language feedback is to provide enriched information to guide the model in revising its response. We can adjust the user preference $\bm{z}$ to generate outputs aligned to different preferences, e.g., detailed or humorous responses.

\begin{algorithm}[t]
\caption{\emph{PRS}}
\label{algo:prs}
\begin{algorithmic}[1]
\State \textbf{Input:} Input prompt $\bm{x}$; preference $\bm{z}$; model $\pi_\theta$; reward model $R$; number of layers $d$; total samples $N$ to generate
\State \textbf{Initialize:} Layer width $w = \left \lfloor \frac{N}{d} \right \rfloor$

\State $\mathcal{Y} \leftarrow \emptyset$
\For{$l = 0$ \textbf{to} $d-1$}
    \State Select $\bm{y}^*$ with the highest score from $\mathcal{Y}$ or set $\bm{y}^*$ to None if $\mathcal{Y}$ is $\emptyset$
    \State $\bm{f} \sim \pi_\theta(\cdot | \bm{x}, \bm{z}, \bm{y}^*)$ if $\bm{y}^*$ is not None else None
    \For{$i = 1$ \textbf{to} $w$}
        \State Sample $\bm{y}_i \sim \pi_\theta(\cdot|\bm{x}, \bm{z}, \bm{y}^*, \bm{f})$
        \State Add $\bm{y}_i$ to $\mathcal{Y}$
    \EndFor
    \State Compute $R(\bm{x}, \bm{z}, \bm{y})$ for newly generated samples in $\mathcal{Y}$
    
\EndFor
\State \textbf{Output:} The best final response $\bm{y}^*$
\end{algorithmic}
\end{algorithm}

\noindent{\textbf{Tree-Based Generation.}} \ For each input, we sample $N$ responses for further selection. However, as Eq.~\ref{eq:pres} indicates, various components (i.e., $\bm{z}$, $\bm{y}_0$, $\bm{f}$) control the generation, causing difficulty in efficient generation. To overcome this issue, we propose tree-based generation (Fig.~\ref{fig:pres}b), which utilizes an iterative exploration and exploitation process:
\begin{enumerate}[itemsep=-5pt, topsep=0pt, leftmargin=0.5cm]
    \item First, the model randomly samples $N_0$ initial responses $\mathcal{Y}_0$ from $\pi_\theta(\bm{y}_0|\bm{x}, \bm{z})$, and the reward model $R(\bm{x}, \bm{z}, \bm{y})$ generates rewards for the samples. The response $\bm{y}_0^*$ with the highest reward is selected for further exploration. 
    
    \item Then the model generates language feedback $\bm{f}$ for $\bm{y}_0^*$, i.e., $\bm{f} \sim \pi_\theta(\bm{f}|\bm{x}, \bm{z}, \bm{y}_0^*)$, which is the suggestion to further modify $\bm{y}_0^*$ to be more in line with the preference $\bm{z}$~(see the example prompt in Fig.~\ref{fig:prompt_feedback_gen}). 

    \item The model generates another set of $N_1 = N - N_0$ refinements $\mathcal{Y}_1$ from $\pi_\theta(\bm{y}_1|\bm{x}, \bm{z}, \bm{y}_0^*, \bm{f})$, where $N$ is the total number of samples per prompt. It aims to adjust the generation towards even better rewards~(see the prompt of Fig.~\ref{fig:prompt_refine_w_feedback}).

    \item We combine $\mathcal{Y}_0$ and $\mathcal{Y}_1$ into $\mathcal{Y}$ that has $N$ samples for the input $\bm{x}$.
    
    \item (Optional) In layer $l$, suppose we have samples $\mathcal{Y}^{(l-1)} = \mathcal{Y}_0 \cup \cdots \cup \mathcal{Y}_{l-1}$ until layer $l-1$. We further sample refinements $\mathcal{Y}_l$ with a--c steps:
    \begin{subequations}
    \vspace{-5pt}
    \renewcommand{\theequation}{\theparentequation\alph{equation}}  
    \begin{align}
        \bm{y}^* &\gets {\arg\max}_{\bm{y}_i \in \mathcal{Y}^{(l-1)}} R(\bm{x}, \bm{z}, \bm{y}_i) \\
        \bm{f} &\sim p(\cdot | \bm{x}, \bm{z}, \bm{y}^*) \\
        \mathcal{Y}_l &\sim p(\cdot | \bm{x}, \bm{z}, \bm{y}^*, \bm{f})
    \end{align}
    \label{eq:go_deep}
    \end{subequations}
    Eq.~\ref{eq:go_deep}a identifies the optimal response from all already generated responses (i.e., exploitation), followed by refinements (exploration).
\end{enumerate}
We present the pseudocode of \emph{PRS} in Algorithm~\ref{algo:prs}. It is worth noting that \emph{PRS} is also functional when the preference $\bm{z}$ is not provided as input. Additionally, feedback can be omitted during the generation of refinements. In Algorithm~\ref{algo:prs}, the number of samples generated for each layer is set to be the same. However, in practice, other hyper-parameters can be used.

\noindent{\textbf{Reward Estimation.}} \ 
In \emph{PRS}, the reward for a response is calculated using the formula \( R(\bm{x}, \bm{z}, \bm{y}) \), where \(\bm{z}\) specifies the preference for aspects to be focused on when assessing the response. However, if the specified preference \(\bm{z}\) aligns with the implicit preference already incorporated into the reward model, the formula can be simplified to \( R(\bm{x}, \bm{y}) \). In this case, the reward model automatically evaluates the response based on its built-in preference, without the need for \(\bm{z}\). To achieve high rewards, it is crucial to understand and articulate the internal preference of the reward model.

\subsection{Alignment for Preference-Controlled Text Generation}
Here, we study the task of preference-controlled text generation. We train the model to produce responses aligned with the input-specified preference, i.e., $\bm{y}(\bm{z}) \sim \pi_\theta(\bm{y} | \bm{x}, \bm{z})$. We adopt offline RL in $\S$~\ref{sec:preliminary_offline_RL} for training, which repeats iterations of data generation and model re-training. 

As indicated by Eq.~\ref{eq:pres}, adjusting the preference $p(\bm{z})$ can generate diverse outputs, each tailored to a specific preference. Without loss of generality, we do not focus on one specific personalized preference. Instead, we consider diverse preferences. We annotate diverse preferences to ensure each input question is associated with a different preference from others. As exemplified by Table~\ref{tab:preference}, the task of instruction following has diverse personalized preferences and for document summarization, the keywords vary for different documents. 

Algorithm~\ref{code:pres} in the Appendix is the pseudocode for training. Specifically, we conduct $K$ iterations of offline RL training. In each iteration $k$, we have an unlabeled set $\mathcal{U}_k = \{( \bm{x}, \bm{z} ) \}$ and we initialize the training set $\mathcal{D}_k$ to $\emptyset$. For each data point $(\bm{x}, \bm{z}) \in \mathcal{U}_k$, we sample $N$ responses in total. We first generate $N_0$ initial responses denoted as $\mathcal{Y}_0$ and then $N_1 = N - N_0$ refinements denoted as $\mathcal{Y}_1$. We use a reward model to select high-quality data for training. To enhance tree-based generation, we aim to optimize the following two components: 

\textbf{$\bullet$ $\pi_\theta(\bm{y} | \bm{x}, \bm{z})$:} It trains the policy to generate responses aligned with input preferences. 
We use the reward model to identify the response $\bm{y}^*$ with the highest reward from $\mathcal{Y}_0 \cup \mathcal{Y}_1$, and we add the data of $( \bm{x}, \bm{z}, \bm{y}^* )$ to the training set $\mathcal{D}_k$. 

\begin{table}[t]
\centering
\resizebox{\linewidth}{!}{%
\begin{tabular}{p{9cm}}
\toprule
\textit{\textbf{Task 1: Instruction following}}                                                                                                                                                                       \\ \midrule
{\ul \textit{Common Preference}}                                                                                                                                                                                         \\
I prefer responses that are informative, precise, creative, detailed, relevant, and in-depth.                                                                                                                 \\ \midrule

{\ul \textit{Personalized Preferences}}                                                                                                                                                                                   \\
$[1]$ I prefer the model to provide a concise and accurate answer without any unnecessary details or explanations.                                                                                                  \\
$[2]$ I prefer clear and well-organized responses that provide step-by-step instructions or explanations. Additionally, I appreciate when the response includes code snippets or examples for better understanding. \\
...            \\ \midrule
\textit{\textbf{Task 2: Keyword-focused summarization}}                                                                                                                                                               \\ \midrule
I prefer a response that is strictly within 3 sentences, focusing on the keywords of \{\emph{\underline{specify three keywords here}}\}.       \\ \bottomrule                                                                                                         
\end{tabular}%
}
    \caption{
    The explicit preferences used for response optimization. They are added after the input question or document. For instruction following, we evaluate common and personalized preferences. 
    }
    \label{tab:preference}
\end{table}

\textbf{$\bullet$ $\pi_\theta(\bm{y} | \bm{x}, \bm{z}, \bm{y}_0, \bm{f})$:} To improve the model's refinement ability, we construct improving pairs from $\mathcal{Y}_0$ and $\mathcal{Y}_1$. 
We only keep samples from $\mathcal{Y}_1$ that are refined based on the response $\bm{y}^*_0$ if their rewards exceed $\bm{y}^*_0$.
The set of improving pairs is formalized as:
\begin{equation}
    \begin{aligned}
        \mathcal{Q} = \Big\{  ( \bm{x}, \bm{z}, \bm{y}^*_0, \bm{f}, \bm{y}_1 ) \ | &  \\  R(\bm{x},\bm{z},\bm{y}_1) > R(\bm{x}, & \bm{z},\bm{y}^*_0), \forall \bm{y}_1 \in \mathcal{Y}_1  \Big\}
    \end{aligned}
\end{equation}
In our setting, if $\mathcal{Q}$ is not empty, we add the improving data of $( \bm{x}, \bm{z}, \bm{y}^*_0, \bm{f}, \bm{y}^*_1 )$ into the training set $\mathcal{D}_k$, where $\bm{y}_1^*$ is the response with the highest reward from $\mathcal{Y}_1$. This is the same idea as best-of-$N$ sampling, to maximize the response's reward after the model's refinement. 

After generating data from $\mathcal{U}_k$, we combine the generated training data up to iteration $k$, i.e., $\mathcal{D} = \mathcal{D}_1 \cup \cdots \cup \mathcal{D}_k$. Then we re-train the policy with the following objective, which refers to the NLL loss in Eq.~\ref{eq:nll}:
\begin{align}
      & \mathcal{L}(\theta)  = 
        \mathbb{E}_{( \bm{x}, \bm{y} ) \sim \mathcal{D}_0} \big[ - \log \pi_{\theta}(\bm{y}|\bm{x}) \big] \notag
      \\ & + \mathbb{E}_{( \bm{x}, \bm{z}, \bm{y}^* ) \sim \mathcal{D}} \big[ - \log \pi_{\theta}(\bm{y}^*|\bm{x}, \bm{z}) \big] \notag  \\ &  + \mathbb{E}_{( \bm{x}, \bm{z}, \bm{y}^*_0, \bm{f}, \bm{y}^*_1 ) \sim \mathcal{D}} \big[  - \log \pi_{\theta}(\bm{y}^*_1|\bm{x}, \bm{z}, \bm{y}^*_0, \bm{f}) \big] 
    \label{eq:re-train}
\end{align}
where the labeled training data $\mathcal{D}_0$ is also included. After $K$ iterations of RL training, we obtain the model $\pi_\theta (\bm{y}|\bm{x}, \bm{z})$ that can generate the response $\bm{y}$ aligned to the preference $\bm{z}$.

\begin{figure*}[t]
\centering
    \begin{minipage}{0.48\textwidth}
        \centering
        \includegraphics[width=\linewidth]{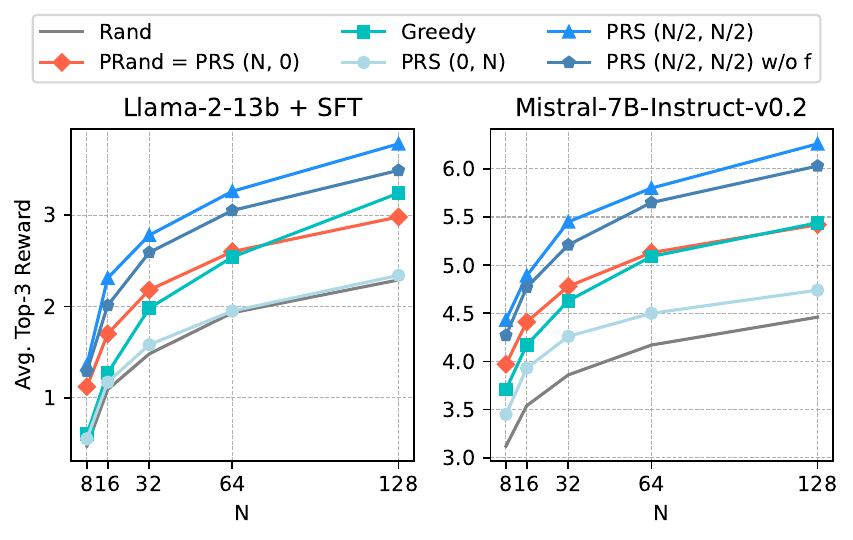}
    \end{minipage}
    \begin{minipage}{0.23\textwidth}
        \centering
        \includegraphics[width=\linewidth]{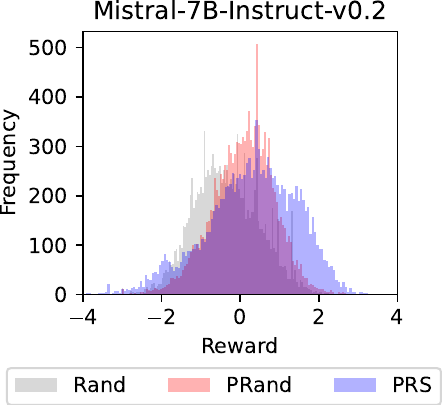}
    \end{minipage}
    \begin{minipage}{0.27\textwidth}
        \centering
        \includegraphics[width=\linewidth]{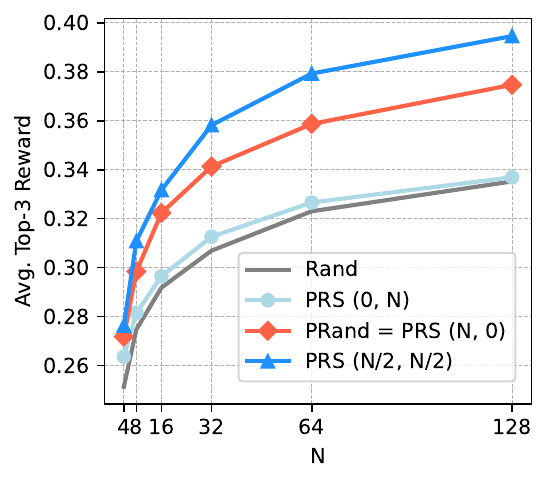}
    \end{minipage}
    \vspace{-12pt}
    \caption{Comparing sampling methods. \textbf{Left:} We study the common preference and use the description of Table~\ref{tab:preference} to generate detailed and in-depth responses. With 100 random prompts from Alpaca-GPT4, each method samples $N$ responses per prompt~(i.e., 8, 16, 32, 64, or 128). The top three highest rewards are averaged for each prompt, leading to an overall average score for the entire evaluation set. The full results of 9 policy models are shown in Fig.~\ref{fig:full-compare-sampling}. 
    \textbf{Middle:} The distribution of rewards with $N=128$, where \emph{PRS} is \emph{PRS $(N/2, N/2)$}. 
    \textbf{Right:} Summarization results on 100 random documents from CNN / Daily Mail. The policy model is Llama-2-13b+SFT.}
    \label{fig:compare-sampling}
    \vspace{-10pt}
\end{figure*}

\begin{table}[t]
\centering
\large
\resizebox{\columnwidth}{!}{%
\begin{tabular}{rccrc}
\toprule
\multicolumn{1}{l}{}                             & \multicolumn{2}{c}{\textbf{AlpacaEval v2.0}} & \multicolumn{2}{c}{\textbf{Arena-Hard v0.1}}                    \\
\textbf{Method}                                  & \textbf{LC WR}        & \textbf{WR}          & \textbf{Method}                                & \textbf{WR} \\ \midrule
\cellcolor[HTML]{F3F3F3}Mis-7B-Inst-v0.2 & 17.10                  & 14.70                 & \cellcolor[HTML]{F3F3F3}mis-large-2407     & 70.4           \\
Rand (Bo-16)                                     & 23.90                  & 19.86                & Rand (Bo-16)                                   & 77.0             \\
\textit{PRS (Bo-16)}                             & \textbf{27.19}        & \textbf{19.87}       & \textit{PRS (Bo-16)}                           & \textbf{79.3}  \\
Rand (Bo-32)                                     & 24.85                 & \textbf{20.61}       & Rand (Bo-32)                                   & 79.1           \\
\textit{PRS (Bo-32)}                             & \textbf{27.17}        & 20.03                & \textit{PRS (Bo-32)}                           & \textbf{80.3}  \\ \midrule
\cellcolor[HTML]{F3F3F3}Lla-3-8b-inst      & 22.90                  & 22.60                 & \cellcolor[HTML]{F3F3F3}Lla-3.1-70b-inst & 55.7           \\
Rand (Bo-16)                                     & 31.00                    & 28.75                & Rand (Bo-16)                                   & 69.5           \\
\textit{PRS (Bo-16)}                             & \textbf{35.05}        & \textbf{31.92}       & \textit{PRS (Bo-16)}                           & \textbf{69.8}  \\
Rand (Bo-32)                                     & 32.94                 & 30.43                & Rand (Bo-32)                                   & 68.2           \\
\textit{PRS (Bo-32)}                             & \textbf{36.70}         & \textbf{33.46}       & \textit{PRS (Bo-32)}                           & \textbf{72.2}  \\ \midrule
\cellcolor[HTML]{F3F3F3}Gemma-2-9b-it            & 48.61                 & 37.07                & \cellcolor[HTML]{F3F3F3}qwen2-72b-inst     & 46.9           \\ 
Rand (Bo-16)                                     & 55.07                 & \textbf{44.51}       & Rand (Bo-16)                                   & 61.9           \\
\textit{PRS (Bo-16)}                             & \textbf{58.40}         & 43.86                & \textit{PRS (Bo-16)}                           & \textbf{62.1}  \\
Rand (Bo-32)                                     & 57.61                 & 45.10                 & Rand (Bo-32)                                   & 63.9           \\
\textit{PRS (Bo-32)}                             & \textbf{59.85}        & \textbf{46.41}       & \textit{PRS (Bo-32)}                           & \textbf{65.4}  \\ \bottomrule
\end{tabular}%
}
\caption{Results of best-of-$N$~(Bo-$N$) sampling on AlpacaEval and Arena-Hard benchmarks, compared to the results of one-pass inference. We use ArmoRM-Llama3-8B-v0.1 as the reward model. Each prompt samples $N$ responses using repeated random sampling or \emph{PRS} and the best response with the highest reward is kept for evaluation. Here, \emph{PRS} does not include preference in the input, and feedback is not generated during refinement. \emph{PRS} uses the version of \emph{PRS}~($N/2$, $N/2$). The higher score between \emph{PRS} and Rand is highlighted in bold. LC WR is the abbreviation for length-controlled win rate.}
\label{tab:bon_benchmark}
\end{table}

\section{Experiments}
\noindent{\textbf{Dataset.}} To align models for preference-controlled text generation, i.e., instruction following and keyword-focused document summarization, we used the following dataset for supervised fine-tuning~(SFT) and RL training:

$\bullet$ \textbf{Instruction Following.} \ For SFT data, from the widely used dataset ShareGPT\footnote{\url{https://sharegpt.com/}}, we randomly sample 10k conversations with a maximum of three rounds for each conversation, resulting in 21,934 labeled data points in total. Prompts from Alpaca-GPT4~\cite{peng2023instruction} are used for RL training.

$\bullet$ \textbf{Summarization.} \ We use the same SFT data from ShareGPT for instruction tuning. We further sample 2,500 documents from CNN / DailyMail~\cite{DBLP:conf/acl/SeeLM17} for RL training. 

\noindent{\textbf{Preference Annotation.}} \ We first need to annotate the preferences for the unlabeled prompts. We show some sample preferences in Table~\ref{tab:preference}. 

$\bullet$ \textbf{Instruction Following.} The Alpaca-GPT4 dataset initially does not include user preferences, so we use GPT-3.5-turbo to automate the generation of preferences by pairing each prompt with a random profession from a list of 222 professions. This method aims to mirror personalized preferences across various professions, thereby enriching dataset diversity. For details on this process and examples, see Appendix~\ref{appendix:preference_gen}. 

$\bullet$ \textbf{Summarization.} To get the input preference keywords, we prompt GPT-3.5-turbo to extract three keywords from the ground-truth summary.

\noindent{\textbf{Benchmarks.}} For instruction following, we use AlpacaEval~\cite{alpaca_eval} with 805 test samples and Arena-Hard~\cite{li2024crowdsourced} with 500 test samples. For summarization, we further sample 1k data from CNN / DailyMail as the test set.

\noindent{\textbf{Reward Model.}} \ For instruction following, we use UltraRM-13B~\cite{cui2023ultrafeedback}, a Llama-2-13B model tuned with GPT-4 preference feedback. It achieved SOTA results on multiple public preference test sets, including Anthropic Helpful~\cite{bai2022training}. 
For summarization, since we lack a reward model, we simulate rewards by comparing summaries to the ground truth using average F1 scores from Rouge-1, -2, and -L~\cite{lin2004rouge}. Lastly, we use ArmoRM-Llama3-8B-v0.1~\cite{wang2024interpretable} for best-of-$N$ sampling on AlpacaEval and Arena-Hard.

\noindent{\textbf{Baselines.}} \ We compare various sampling methods with \emph{PRS}:
\begin{itemize}[leftmargin=0.4cm, itemsep=-4pt, topsep=2pt]
    \item \textbf{Rand} is repeated random sampling conditioned on the input $\bm{x}$ using $\pi_{\theta} (\bm{y} | \bm{x})$. 
    
    \item \textbf{PRand} adds an explicit preference $\bm{z}$ to the input $\bm{x}$, i.e., $\bm{y} \sim \pi_\theta(\bm{y}|\bm{x}, \bm{z})$, for random sampling. 
    
    \item \textbf{Greedy} utilizes a greedy algorithm, where we improve the method from \citet{madaan2024self} which iteratively refines the last response. Specifically, the enhanced baseline starts by sampling an initial response with \(\pi_\theta(\bm{y}_0|\bm{x},\bm{z})\). It uses a reward model to continually update the highest-reward response \(\bm{y}^*\) with \(\pi_\theta(\bm{y} | \bm{x}, \bm{z}, \bm{y}^*)\). During each revision round, if a new response \(\bm{y}\) achieves a higher reward, it becomes \(\bm{y}^*\). 
\end{itemize}
We use temperature sampling during response generation.

\subsection{Comparison of Sampling Methods}\label{sec:compare_sampling}
We first compare different sampling methods for data generation. We expect a good sampling method to obtain a training set with a higher reward. Here, we only consider two layers for the tree-based generation in \emph{PRS}. 
Since the \textbf{\emph{PRS $(N_0, N_1)$}} method is affected by the hyper-parameters $N_0$ and $N_1$, we adjust them to examine their impact:
\begin{itemize}[leftmargin=0.4cm, itemsep=-4pt, topsep=2pt]
    \item \textbf{\emph{PRS $(0, N)$}} samples one response $\bm{y}_0$, generates feedback $\bm{f}$, and then samples $N$ refinements. It neglects the exploration for $\bm{y}_0$.
    
    \item \textbf{\emph{PRS $(N, 0)$}} samples $N$ responses of $\bm{y}_0$ without refinement, which neglects the exploration of $\bm{y}_1$. This is precisely the PRand baseline.

    \item \textbf{\emph{PRS $(N/2, N/2)$}} balances exploring $\bm{y}_0$ and $\bm{y}_1$. 

    \item \textbf{\emph{PRS $(N/2, N/2)$ w/o f}} omits generating language feedback $\bm{f}$ during refinement and instead uses $\pi_\theta(\bm{y}_1 | \bm{x}, \bm{z}, \bm{y}^*_0)$. The goal is to assess the impact of language feedback.
\end{itemize}

\noindent{\textbf{Policy Models.}} \ We use the model tuned on the SFT data from ShareGPT named \emph{Llama-2-13B + SFT} to sample responses. We also test multiple open-source instruction-following models such as those tuned on Mistral-7B~\cite{jiang2023mistral} and Llama-2-13b~\cite{touvron2023llama}.

\begin{table}[t]
    \centering
    \resizebox{\linewidth}{!}{%
    \begin{tabular}{lccc}
    \toprule
      Test Set Size  & \multicolumn{1}{c}{805}   & \multicolumn{1}{c}{200}    & \multicolumn{1}{c}{200}  \\ \midrule
    Annotator        & \multicolumn{1}{c}{{GPT4}}      & \multicolumn{1}{c}{{GPT4}}    & \multicolumn{1}{c}{{GPT4}} \\ 
    Baseline Methods & \multicolumn{1}{c}{{davinci-3}} & \multicolumn{1}{c}{{ChatGPT}} & \multicolumn{1}{c}{{GPT4}} \\  
                                     & \multicolumn{1}{c}{\% Win} & \multicolumn{1}{c}{\% Win} & \multicolumn{1}{c}{\% Win} \\  \midrule
    GPT 3.5 Turbo 0301                      & \multicolumn{1}{c}{89.37}  & \multicolumn{1}{c}{50.00}     & \multicolumn{1}{c}{-}      \\ 
    UltraLM 13B V2.0 (PPO) & \multicolumn{1}{c}{86.30} & \multicolumn{1}{c}{-} & \multicolumn{1}{c}{-} \\
    LLaMA2 Chat 13B (PPO)    & \multicolumn{1}{c}{81.09}  & \multicolumn{1}{c}{-}      & \multicolumn{1}{c}{-}      \\
    Tulu 2 13B (SFT) & \multicolumn{1}{c}{78.90} & \multicolumn{1}{c}{-} & \multicolumn{1}{c}{-} \\ \midrule
    SFT + p                                 & \multicolumn{1}{c}{80.64}  & \multicolumn{1}{c}{53.27}  & \multicolumn{1}{c}{17.59}  \\ 
    Base + p                                & \multicolumn{1}{c}{79.61}  & \multicolumn{1}{c}{51.26}  & \multicolumn{1}{c}{22.11}  \\ 
    \multicolumn{4}{l}{\emph{\underline{Offline RL training on Base with various sampling methods}}} \\ 
    Rand + p                                & \multicolumn{1}{c}{82.60}   & \multicolumn{1}{c}{59.05}  & \multicolumn{1}{c}{30.81}  \\ 
    Rand      & \multicolumn{1}{c}{80.40}   & \multicolumn{1}{c}{49.75}  & \multicolumn{1}{c}{23.37}  \\ 
    PRand                                   & \multicolumn{1}{c}{85.07}  & \multicolumn{1}{c}{64.32}  & \multicolumn{1}{c}{39.20}   \\ 
    \textbf{\emph{PRS}}                                  & \multicolumn{1}{c}{\textbf{86.89}}  & \multicolumn{1}{c}{\textbf{72.36}}  & \multicolumn{1}{c}{\textbf{43.22}}  \\ 
    \bottomrule
    \end{tabular}%
    }
    \captionof{table}{Results of AlpacaEval v1.0. We use the common preference in Table~\ref{tab:preference} to control our models to generate responses. ``+ p'' adds preference in the input during testing. \emph{SFT} uses all available labeled data of ShareGPT and Alpaca-GPT4 for supervised fine-tuning. \emph{Base} is the model tuned using ShareGPT data. To reduce the cost of calling GPT-4, we downsampled the test set for ChatGPT and GPT-4 baseline. We also show existing models tuned from Llama-2-13B for comparison, but they are fine-tuned with full parameters and different training data.}
    \label{tab:result_AlpacaEval}
\end{table}

\begin{table}[t]
    \centering
    \resizebox{\linewidth}{!}{%
    \begin{tabular}{lcccc}
    \toprule
    \textbf{}         & \textbf{R-1}             & \textbf{R-2}             & \textbf{R-L}             & \textbf{Avg.}            \\ \midrule
    LLaMA2 Chat 13B   & 32.93           & 10.70           & 29.29           & 24.31          \\
    Mistral 7B v0.2   & 34.98           & 11.27           & 31.38           & 25.88          \\
    Tulu 2+DPO 13B    & 36.64           & 12.93           & 33.34           & 27.64          \\
    Vicuna 13B V1.5   & 37.12           & 13.26           & 33.71           & 28.03          \\
    \midrule
    Base w/o keywords  & 30.15           & 10.35           & 27.89           & 22.80          \\
    Base + p             & 35.46           & 12.56           & 32.37           & 26.80          \\
    \multicolumn{5}{l}{{\ul \textit{RL training on un-tuned Llama-2-13B}}}            \\
    PRand          & 37.39$^\dag$           & 13.71$^\dag$           & 33.96$^\dag$           & 28.35$^\dag$          \\
    \textbf{\textit{PRS}}  & \textbf{38.20$^*$}  & \textbf{14.16$^*$} & \textbf{34.70$^*$}  & \textbf{29.02$^*$} \\
    \multicolumn{5}{l}{{\ul \textit{Continual RL training on Base}}}                           \\
    PRand          & 37.50$^\dag$            & 13.78$^\dag$           & 34.12$^\dag$           & 28.47$^\dag$          \\
    
    \textbf{\textit{PRS}}  & \textbf{38.15$^*$} & \textbf{14.16$^*$} & \textbf{34.65$^*$} & \textbf{28.99$^*$} \\ \bottomrule
    \end{tabular}%
    }
    \captionof{table}{Summarization results on CNN / Daily Mail, adding input keywords except for the ``Base w/o keywords'' condition. We report average Rouge-1, Rouge-2, and Rouge-L F1 scores with 5 runs. $*$ indicates \emph{PRS} outperforms PRand significantly ($p < 0.01$), and $\dag$ indicates PRand outperforms Vicuna 13B V1.5 ($p < 0.01$).
    }
    \label{tab:summarization}
\end{table}

\begin{figure*}[t]
\centering
    \begin{minipage}{0.25\textwidth}
        \centering
        \includegraphics[width=\linewidth]{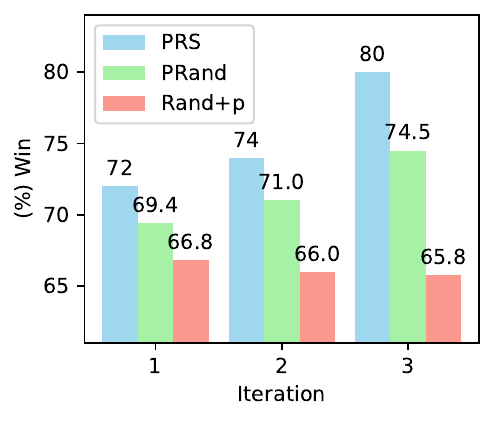}
        \vspace{-22pt}
        \caption{Offline RL training: Win rates for \emph{PRS}, PRand, and Rand + p vs. Base + p, evaluated using GPT-4 on a 200-sample AlpacaEval. ``+ p'' adds common preference in the input.}
        \label{fig:result_win_vs_base}
    \end{minipage}
    \hspace{1pt}
    \begin{minipage}{0.48\textwidth}
        \centering
        \includegraphics[width=\linewidth]{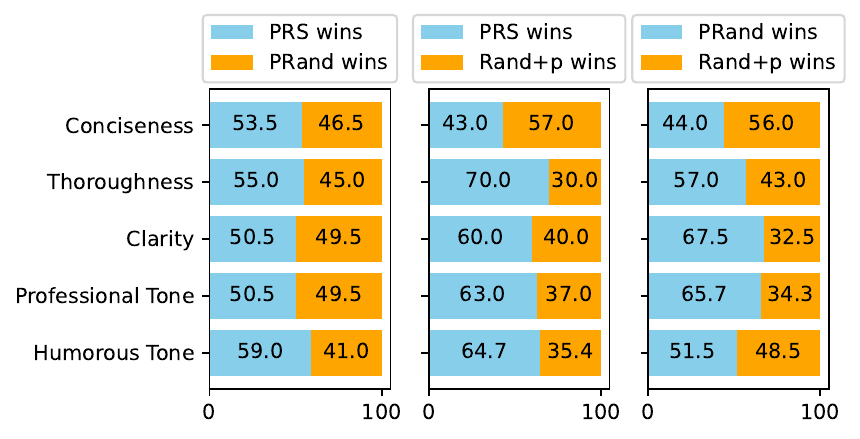} 
        \vspace{-18pt}
        \caption{Preference Adaptation: We define five preference categories and evaluate each category using 100 AlpcaEval test cases. For each category, we customize the prompt (100 test samples) by appending the corresponding preference, evaluating with GPT-4, and recording win rates (\%) when comparing two models.}
        \label{fig:win_preference_adapt}
    \end{minipage}
    \hspace{1pt}
    \begin{minipage}{0.24\textwidth}
        \centering
        \Large
        \resizebox{\linewidth}{!}{%
        \begin{tabular}{lc}
        \toprule
        \textbf{ToxiGen}       & \% Toxic ($\downarrow$)     \\ \midrule
        GPT-4-0613             & 0.6          \\
        GPT-3.5-turbo-0613     & 0.5          \\
        GPT-3.5-turbo-0301     & 27.7         \\
        Zephyr 7B Beta         & 64.0         \\
        Xwin-LM v0.1 70B       & 12.7         \\
        Tulu 2+DPO 13B         & 1.1          \\ \midrule
        Rand                   & 3.9          \\
        Rand + p               & 0.3          \\
        PRand                  & 0.2 \\
        \textit{\textbf{PRS}} & 0.2 \\ \bottomrule
        \end{tabular}%
        }
        \captionof{table}{Toxicity reduction. We append a preference indicating a safe response in the input for Rand + p, PRand, and \emph{PRS}.}
    \label{tab:toxic}
    \end{minipage}
    \vspace{-10pt}
\end{figure*}

\noindent{\textbf{Preference $\bm{z}$.}} \ For instruction following, we aim to evaluate the common preference (as shown in Table~\ref{tab:preference}) that favors comprehensive and detailed responses. As shown by \citet{sun2023principle}, a more detailed response would improve the performance on benchmarks such as AlpcaEval~\cite{alpaca_eval}. 
Since the reward model UltraRM-13B that we use internally includes such preferences, we compute $R(\bm{x},\bm{y})$ without explicitly specifying $\bm{z}$.

\noindent{\textbf{Results.}} \ From the results shown in Fig.~\ref{fig:compare-sampling}, \emph{PRS} generates data with higher rewards than Rand and PRand, and as $N$ increases, the performance gap becomes larger. The setting $(N/2, N/2)$ is much better than $(0, N)$ and $(N, 0)$, showing that a good balance of exploration is important. Fig.~\ref{fig:compare-sampling}~(middle) shows that \emph{PRS} produces a normal distribution with higher mean and variance than PRand and Rand, indicating a broader exploration and higher reward acquisition in the sampling space. From the full results shown in Fig.~\ref{fig:full-compare-sampling}, language feedback shows mixed results: some models improve, while others do not. However, language feedback increases transparency and both versions still outperform other baselines.

PRand is substantially better than Rand, since PRand adds explicit preference in the input. It demonstrates that preference is effective in guiding the generation of better-aligned responses. For summarization, specifying the keywords would aid the model to concentrate on the key information of the document. The greedy algorithm, revising based on the current best response, often underperforms compared to \emph{PRS}. Its main limitation is poor response exploration. In contrast, \emph{PRS $(N/2, N/2)$} excels by thoroughly exploring both initial and subsequent responses.

We further investigate best-of-$N$ sampling on AlpacaEval v2.0 and Arena-Hard v0.1. The models are evaluated as outlined in Table~\ref{tab:bon_benchmark}. To obtain the reward scores, we utilize the recent state-of-the-art reward model, ArmoRM-Llama3-8B-v0.1. For \emph{PRS}, no preference is specified, and feedback generation is omitted during sampling to support more general use cases. We employ two layers in \emph{PRS}, with each layer having a width of $N/2$. As shown in Table~\ref{tab:bon_benchmark}, \emph{PRS} consistently outperforms repeated random sampling, achieving better performance in LC WR on AlpacaEval and WR on Arena-Hard.

\subsection{Offline RL Training}\label{sec:offline_RL}
We conduct offline RL training to align the models to generate responses tailored to input preferences.

\noindent{\textbf{Experimental Settings.}} \ 
We fine-tune the Llama-2-13B model using LoRA~\cite{hu2021lora}, starting with supervised fine-tuning (SFT) using labeled data. For instruction following, we perform 3 iterations of RL training, each involving 10k unique GPT-4 prompts. We adopt best-of-16 sampling, generating 16 responses per prompt, and adding 10k new training data per iteration. We set $N_0 = N_1 = 8$ for \emph{PRS}. For summarization, after the initial SFT, we undertake one RL iteration, sampling 64 summaries per document (2,500 in total), retaining the summary with the highest reward for each document. We set $N_0 = N_1 = 32$ for \emph{PRS}.

\noindent{\textbf{Results.}} \ Results of AlpacaEval and CNN/Daily Mail are reported in Tables~\ref{tab:result_AlpacaEval} and \ref{tab:summarization} respectively.

The model trained by \emph{PRS} outperforms those trained by PRand and Rand. Looking at the rewards of the generated training data shown in Fig.~\ref{fig:train_reward_prand_pres} in the Appendix, \emph{PRS} exhibits consistently higher rewards than PRand. It shows that the quality of data generation is key to offline RL. Compared to open-source models, \emph{PRS} outperforms the models tuned by PPO training. In head-to-head comparison shown in Fig.~\ref{fig:result-compare-pres-baselines} in the Appendix, \emph{PRS} outperforms multiple strong open-source models more than 50\% of the time, except for Mistral-7B-v0.2. These promising results highlight the potential of \emph{PRS} for future applications, such as integrating \emph{PRS} with DPO training~\cite{rafailov2024direct} and full-parameter fine-tuning. For summarization, after aligning the model with \emph{PRS}, our model performs the best among existing strong open-source models. 

Preference-controlled optimization during training is important. The method Rand + p involves adding a preference to the input prompt at test time. It effectively enhances performance compared to Rand. However, it does not explicitly optimize the response to the input preference during training compared to PRand, so it underperforms PRand

We further present the results of RL training for each iteration in Fig.~\ref{fig:result_win_vs_base}. Our findings indicate that while using random sampling (Rand) leads to a halt in improvement after just one iteration of RL training, both PRand and \emph{PRS} continue to show improvement across 3 training iterations. The quality of data generated through random sampling can significantly influence the iterative updates made to the model. Since the generated data is of lower quality, it can lead to a degradation in the model's performance. This, in turn, makes it increasingly challenging for the model to generate high-quality data, thereby halting further improvements.

\subsection{Further Analysis}\label{sec:further-analysis}
\noindent{\textbf{Preference Adaptation.}} We further compare \emph{PRS}, PRand, and Rand + p on adaptation to personalized preferences differing from the common preference studied in Fig.~\ref{fig:compare-sampling} (left) and Table~\ref{tab:result_AlpacaEval}. We define five categories as shown in Fig.~\ref{fig:win_preference_adapt} for adaptation and for each category, we create 20 unique expressions using GPT-4. We evaluate them across 100 AlpacaEval test cases. For each category, we randomly sample an expression and append it to the prompt. More details can be found in Appendix~\ref{sec:preference_adaptation}. 

\emph{PRS} outperforms PRand, especially in delivering concise, thorough, and humorous responses. Both models perform similarly in clarity and professional tone. Overall, both \emph{PRS} and PRand surpass Rand + p in effectiveness, showing the benefits of training models to align with user preferences. However, Rand + p excels in conciseness, producing fewer tokens (176.07) compared to \emph{PRS} (199.31). In contrast, for thoroughness, while Rand + p averages 378.99 tokens, PRand and \emph{PRS} provide more thorough responses with 461.34 and 507.81 tokens, respectively.

\noindent{\textbf{Toxicity Reduction.}} \ We further study toxicity reduction as preference adaptation. For each input, we append a safe preference after it, which is randomly sampled from a pool of safe preferences with different expressions~(see Table~\ref{tab:safe_preference}). 
We evaluate ToxiGen~\cite{hartvigsen2022toxigen} and report the results in Table~\ref{tab:toxic}. Compared to Rand and Rand + p, adding a safe preference can substantially reduce the generation of toxic content. PRand and \emph{PRS} achieve comparable performance and both outperform Rand + p. Preference-controlled alignment adapts the LLM to generate safe and harmless content at test time, even without explicit training for safety.

\begin{figure}[t]
    \centering
    \begin{minipage}{0.5\textwidth}
    \begin{minipage}{0.57\linewidth}
    \centering
    \includegraphics[width=\linewidth]{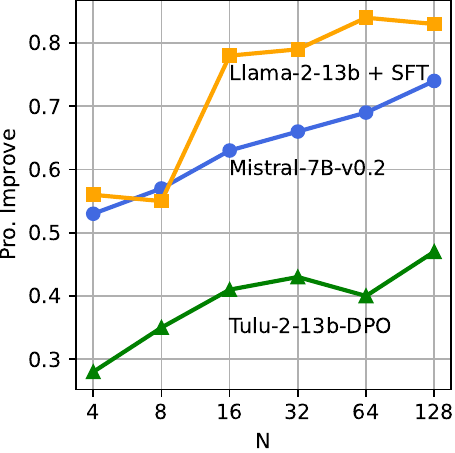}  
\end{minipage}
\begin{minipage}{0.37\linewidth}
    \centering
    \includegraphics[width=\linewidth]{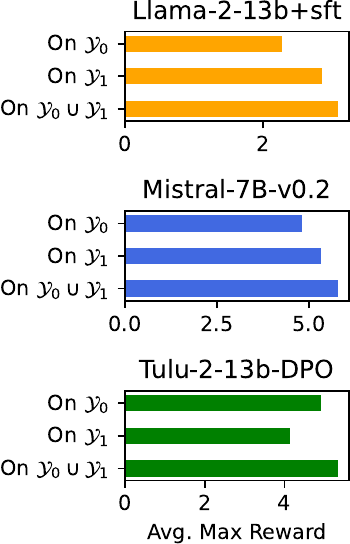} 
\end{minipage}
\caption{Proportion of cases where the top response from \( N_1 \) refinements in \( \mathcal{Y}_1 \) yields a higher reward than the best initial response from \( \mathcal{Y}_0 \). Average maximum rewards for each set and their union are reported ($N$=32).}
\label{fig:pro_improve}
    \end{minipage}
\end{figure}

\noindent{\textbf{Tree-Based Generation.}} \ We analyze tree-based generation in \emph{PRS}, which starts with \(N_0\) initial responses (\(\mathcal{Y}_0\)), and then \(N_1\) refinements (\(\mathcal{Y}_1\)). We evaluate how often refinements improve over the initial response. As shown in Fig.~\ref{fig:pro_improve}, there is variability across models: Tulu-2-13b-DPO improves less than 50\% of the time, while Mistral-7B-v0.2 and Llama-2-13B + SFT perform better. 
Improvement rates generally increase with more samples (\(N\)), indicating that more samples can lead to better outcomes. 
We explore the reward values for $\mathcal{Y}_0$ and $\mathcal{Y}_1$. We find that $\mathcal{Y}_1$ does not consistently offer higher rewards than $\mathcal{Y}_0$, but combining both sets yields higher rewards.

\begin{figure}[t]
\centering
\includegraphics[width=6.5cm]{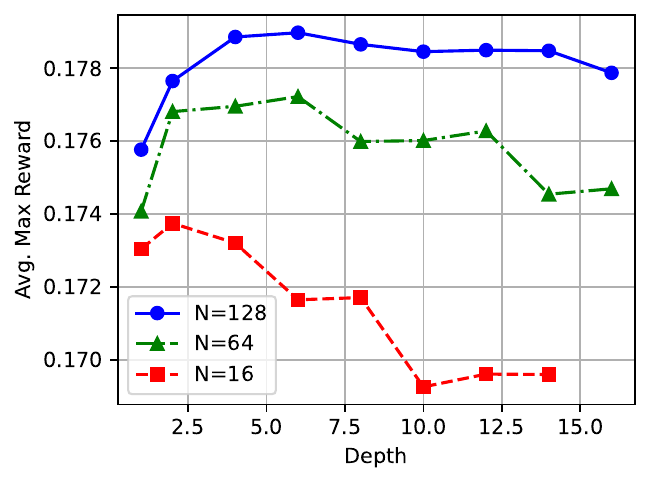} 
\caption{Effects of varying depth and width for \emph{PRS}. We maintain the number of samples $N$ and vary the depth $d$ and the width $w$ calculated by $\left \lfloor \frac{N}{d} \right \rfloor$. The depth starts from 1 to 16. Preference is not included in the input and feedback is not generated. Here, the studied model is Llama-3-8b-instruct and the reward model is ArmoRM-Llama3-8B-v0.1. 100 test samples are randomly selected from AlpacaEval for evaluation.}
\label{fig:go_deeper}
\end{figure}

\noindent{\textbf{Expansion in \emph{PRS}.}} \ Here, we examine how the depth and width impact the performance of tree-based generation in \emph{PRS}. We keep the total number of samples $N$ constant while varying the depth $d$. The width $w$ is then calculated by $\left \lfloor \frac{N}{d} \right \rfloor$. As shown in Fig.~\ref{fig:go_deeper}, our results indicate that for larger $N$, increasing the depth (e.g., to 4) improves performance. However, for smaller values of $N$, such as 16, increasing the depth beyond 2 does not yield further benefits. A larger $N$ results in a greater width, allowing the model to sample more responses at each layer, thereby increasing the likelihood of discovering better responses than those in the previous layers. We further conduct an ablation study in Appendix~\ref{sec:ablation}.

\section{Conclusion}

We introduce \emph{PRS}, an improved sampling method designed to enhance iterative model improvement. In contrast to repeated random sampling, \emph{PRS} enables more efficient generation through a tree-based approach. By allowing the specification of preference in the input, \emph{PRS} optimizes responses to better align language models with diverse user preferences. Our comprehensive evaluation shows that \emph{PRS} consistently generates higher-quality samples. On AlpacaEval and Arena-Hard, \emph{PRS} significantly outperforms random sampling in the best-of-$N$ setting. Additionally, \emph{PRS} excels when applied to iterative offline RL training.

\section{Limitations}
Our approach capitalizes on the model's self-improvement capabilities to aid in data sampling. However, for more challenging tasks, such as reasoning tasks, the model may struggle to enhance its performance autonomously. We have not explored these types of tasks in this work. Further enhancing the model's self-improvement capabilities, particularly for more difficult tasks, can be explored in the future. Our approach may be susceptible to reward hacking, though further research may mitigate its effects.

\section*{Acknowledgements}
This research is supported by the National Research Foundation, Singapore under its AI Singapore Programme (AISG Award No: AISG2-PhD-2021-08-016[T]). We thank Ruochen Xu for his comments on this paper, and the anonymous reviewers for their valuable suggestions.

\bibliography{output}

\appendix

\begin{algorithm*}[!t]
\caption{\textbf{:} \ \emph{PRS} for aligning language models for diverse preferences}
\label{code:pres}
\begin{algorithmic}[1]
\State \textbf{Input:} Labeled training data $\mathcal{D}_0$; $K$ sets of unlabeled data $[\mathcal{U}_1, \cdots, \mathcal{U}_K]$; large language model $\pi_\theta$; reward model $R$; number of samples per prompt $N$; $N_0$.

\State Initialize $\pi_{\theta_0}$ on $\mathcal{D}_0$ using Eq.~\ref{eq:nll}.

\State $\mathcal{D} \gets \emptyset$.

\For{$k = 1$ to $K$}
\State \textcolor{blue}{\# Stage 1: Data Generation}
\State $\mathcal{D}_k \gets \emptyset$.

\ForAll{$( \bm{x}, \bm{z} ) \in \mathcal{U}_k$}
\State \emph{\textcolor{orange}{\# Preference-Guided Reflective Sampling (PRS)}}
\State $\bullet$ Sample $N_0$ responses $\mathcal{Y}_0 \sim \pi_{\theta_{k-1}}(\bm{y}_0 | \bm{x}, \bm{z})$. 

\State Maximize reward $R(\bm{x}, \bm{z}, \bm{y}^i_0)$ over $\mathcal{Y}_0$ to find the optimal $\bm{y}^*_0$.

\State $\bullet$ Generate language feedback $\bm{f} \sim \pi_{\theta_{k-1}}(\bm{f}|\bm{x}, \bm{z}, \bm{y}^*_0)$.

\State $\bullet$ Sample $N_1=N-N_0$ refinements  $\mathcal{Y}_1 \sim \pi_{\theta_{k-1}}(\bm{y}_1 | \bm{x}, \bm{z}, \bm{y}^*_0, \bm{f}) $.

\State Maximize reward $R(\bm{x}, \bm{z}, \bm{y}^i_1)$ over $\mathcal{Y}_1$ to find the optimal $\bm{y}^*_1$.

\If{$R(\bm{x}, \bm{z}, \bm{y}^*_1) > R(\bm{x}, \bm{z}, \bm{y}^*_0)$}
\State Add $( \bm{x}, \bm{z}, \bm{y}^*_1 )$ and $( \bm{x}, \bm{z}, \bm{y}^*_0, \bm{f}, \bm{y}^*_1 )$  into $\mathcal{D}_k$.
\Else
\State Add $( \bm{x}, \bm{z}, \bm{y}^*_0 )$ into $\mathcal{D}_k$.
\EndIf
\EndFor

\State $\mathcal{D} \gets \mathcal{D} \cup \mathcal{D}_k$.

\State \textcolor{red}{\# Stage 2: Model Re-training}
\State Update $\pi_{\theta_k}$ on $\mathcal{D} \cup \mathcal{D}_0$ with Eq.~\ref{eq:re-train}.
\EndFor
\State \textbf{Output:} $\pi_{\theta_K}(\bm{y}|\bm{x},\bm{z})$.
\end{algorithmic}
\end{algorithm*}

\section{Detailed Related Work}\label{app:related_work}
\subsection{Alignment of Large Language Models}

Similar to \emph{PRS},  \citet{DBLP:journals/corr/abs-2212-08073} also leverage the LLM's capacity for reflection to refine model responses. However, our work differs from \citet{DBLP:journals/corr/abs-2212-08073} in several aspects: (a) Most importantly, \citet{DBLP:journals/corr/abs-2212-08073} do not aim to improve data sampling for RLHF, but our work proposes a tree-based framework to enable efficient data generation. (b) \citet{DBLP:journals/corr/abs-2212-08073} only focus on harmless responses, but our work applies to a broader spectrum of preferences. (c) While the preferences added into the input to guide model generation -- introduced in our work -- is similar to the usage of principles proposed by \citet{DBLP:journals/corr/abs-2212-08073}, their approach is limited to modifying responses based on principles rather than integrating these principles into the input prompt to guide the generation of model responses. 

\citet{DBLP:journals/corr/abs-2310-05910} propose to train a reward model that can evaluate responses based on principles, which is similar to our work when using the reward model by adding the extra preference information. However, \citet{DBLP:journals/corr/abs-2310-05910} also overlook the importance of sampling efficiency. Another notable contribution is from \citet{scheurer2023training}, who advocate for training models using language feedback, as opposed to the numerical feedback derived from reward models. Unlike our strategy, which employs the model itself to generate language feedback, they depend on human annotators for this task. Recent work by \citet{feng2023alphazero} aligns with our goal to enhance model sampling exploration. They adopt Monte Carlo tree search~(MCTS) for decoding, utilizing token-level rewards to guide output sampling in instruction-following tasks. In contrast, our approach prioritizes sequence-level rewards for evaluating model responses and employs a tree-based search without extensive inference costs. Furthermore, we incorporate input prompt preferences to direct the generative process, which is another difference from \citet{feng2023alphazero}.

\subsection{Reflective Reasoning of Large Language Models}
Large language models (LLMs) have demonstrated self-reflection capability, critically analyzing their own decisions and providing feedback to enhance their responses~\cite{madaan2024self}. \citet{madaan2024self}  introduce a self-refinement framework that enables LLMs to continuously improve their responses based on self-generated feedback. In contrast, our work introduces an efficient tree-based generation model that optimizes the use of LLMs' reflective abilities more effectively. Further exploring the potential of LLMs' self-reflective capabilities, \citet{shinn2024reflexion} leverage this feature to enable LLMs to learn from language-based feedback and refine their outputs towards more accurate and contextually relevant responses.  In the realm of iterative inference for text generation, \citet{welleck2022generating} propose training a corrector model to refine LLM outputs, utilizing synthetically generated data that fosters gradual improvement. The concept of reflection in LLMs is crucial for advances of AI agents, facilitating their ability to summarize and reflect on outcomes from previous interactions to better plan and execute future actions~\cite{yao2024tree,yao2022react}. 

\subsection{Controlled Instruction Following}
In the era of large language models, there is growing interest in evaluating and enhancing complex instruction following with the outputs controlled by input constraints~\cite{chen2024benchmarking,he2024complex,yao2023collie}. In our work, to improve sampling efficiency, we frame generation as a problem of controlled text generation by treating user preference as the constraint.

\section{Additional Results}

\subsection{Full Results of Data Sampling}
\begin{figure*}[t]
\centering
\includegraphics[width=\textwidth]{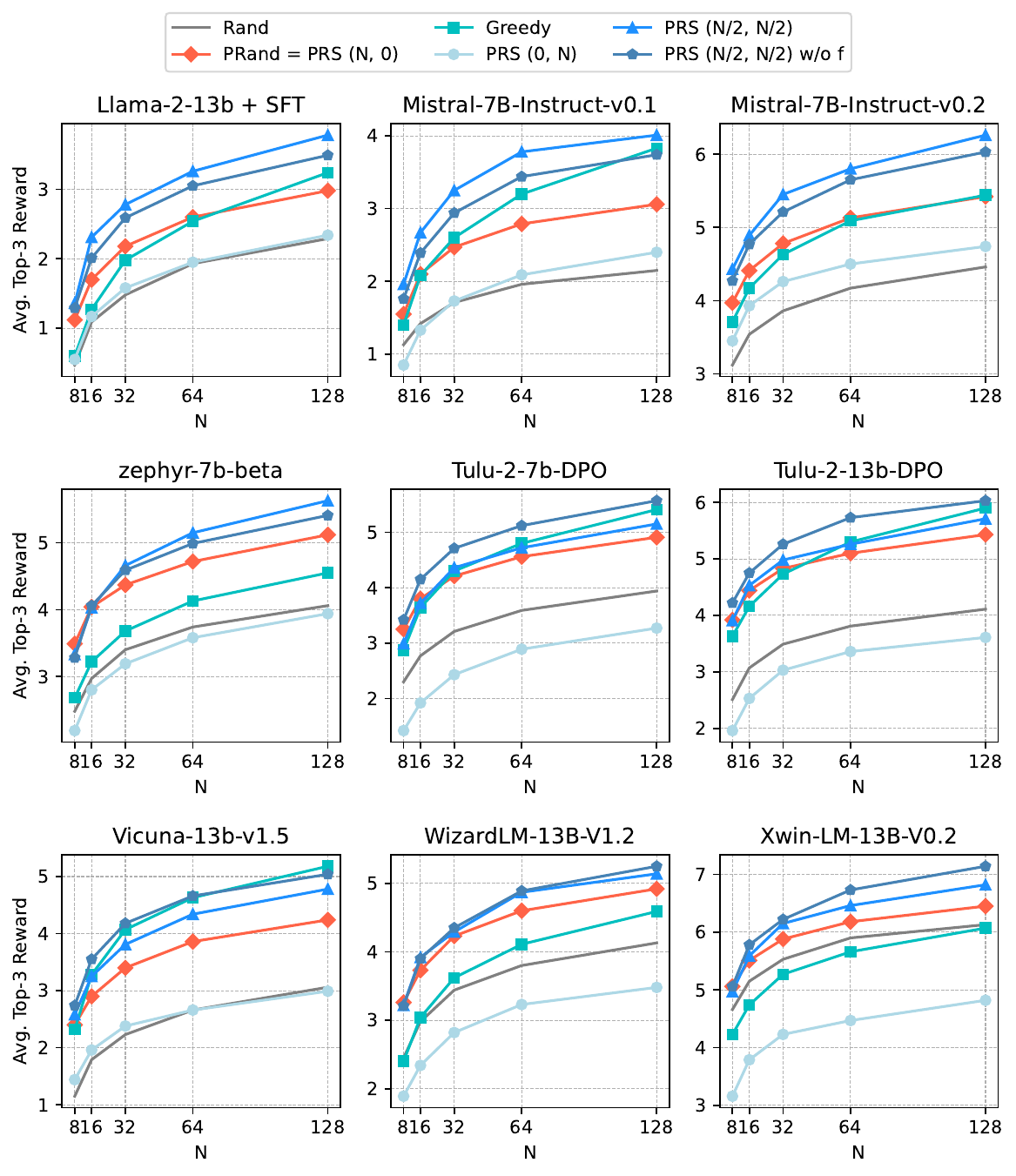} 
\caption{\textbf{Results of data generation for instruction following}: We focus on the common preference and use the description in Table~\ref{tab:preference} to
generate detailed and in-depth responses. 
(a) \textbf{Policy Models}: We use 9 policy models to generate training data, which are Llama-2-13b + SFT, Mistral-7B-Instruct-v0.1, Mistral-7B-Instruct-v0.2, zephyr-7b-beta, Tulu-2-7b-DPO, Tulu-2-13b-DPO, Vicuna-13b-v1.5, WizardLM-13B-V1.2 and Xwin-LM-13B-V0.2. \\
(b) \textbf{Test samples}: We randomly sample 100 prompts from Alpaca-GPT4. \\
(c) \textbf{Setup}: We sample $N$ responses per prompt~(i.e., 8, 16, 32, 64, or 128) using a specific sampling method. We then average the top three rewards for each prompt, leading to an overall average score for the entire evaluation set. We use UltraRM-13B to generate the reward. 
}
\label{fig:full-compare-sampling}
\end{figure*}

We show the full results of data generation on 9 policy models in Fig.~\ref{fig:full-compare-sampling}. 

\subsection{Instruction Following}
\begin{figure*}[t]
     \centering
        \includegraphics[width=0.5\linewidth]{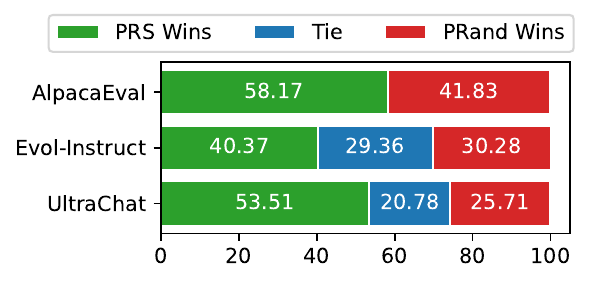} 
        \caption{Head-to-head evaluation of \emph{PRS} and PRand after 3 iterations of RL training. We use GPT-4 as the evaluator.}
        \label{fig:pres_vs_prand}
\end{figure*}

Head-to-head comparison of \emph{PRS} and PRand after 3 iterations of RL training is shown in Fig.~\ref{fig:pres_vs_prand}.

\begin{figure*}[t]
    \centering
    \includegraphics[width=0.4\linewidth]{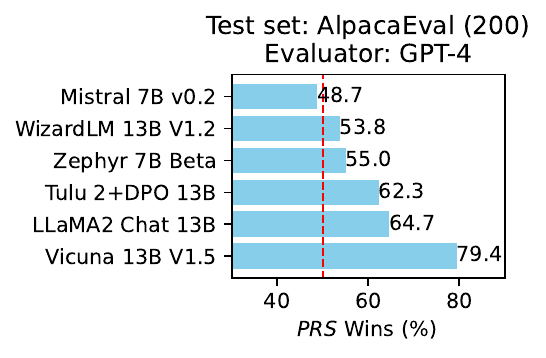} 
        \vspace{-10pt}
        \captionof{figure}{\emph{PRS} vs. open-source models.}
        \label{fig:result-compare-pres-baselines}
\end{figure*}

\begin{table*}[t]
\centering
\small
\resizebox{0.8\linewidth}{!}{%
\begin{tabular}{lp{12cm}}
\toprule
  & \textbf{Safe and Harmless Preferences}                                                                                                 \\ \midrule
1 & I prefer a response that is considerate, safe, and avoids any sensitive or potentially harmful content, without causing any offense or discomfort.            \\ \midrule
2 &
  I prefer a response that is mindful, secure, and steers clear of any sensitive or potentially damaging material, ensuring it doesn't cause offense or discomfort. \\ \midrule
3 & I prefer a response that is thoughtful, safe, and carefully avoids any sensitive or potentially harmful subjects, ensuring no offense or discomfort is caused. \\ \midrule
4 & I prefer a response respectful, cautious, and deliberately evades any delicate or potentially injurious topics, so as to prevent causing any upset or unease.   \\ \midrule
5 &
  I prefer a response that's thoughtful and secure, meticulously avoiding topics that might be sensitive or harmful, thereby ensuring nobody feels offended or uncomfortable. \\ \bottomrule
\end{tabular}%
}
\caption{The safe and harmless preferences used for toxicity reduction.}
\label{tab:safe_preference}
\end{table*}

\begin{figure*}[t]
    \centering
    \includegraphics[width=0.5\linewidth]{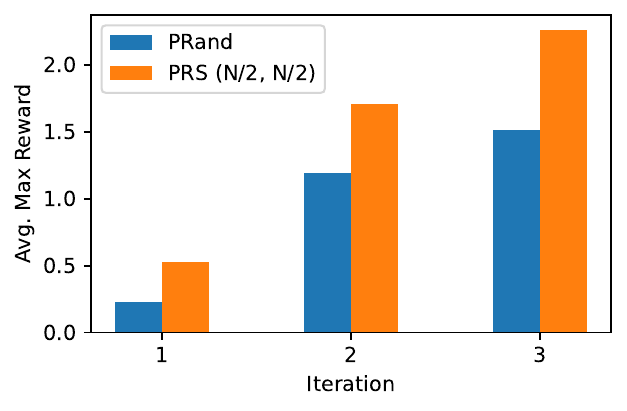}
    \vspace{-10pt}
    \caption{Average rewards of training data for personalized preferences during RL training. 10k prompts from Alpaca-GPT4 are used for sampling, each has a different preference exemplified by Table~\ref{tab:preference}.} 
    \label{fig:train_reward_prand_pres}
\end{figure*}

\subsection{Ablation Study}\label{sec:ablation}
\begin{figure*}[t]
    \centering
    \includegraphics[width=0.35\textwidth]{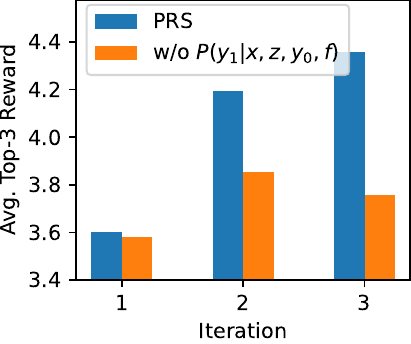}
    \caption{We ablate to exclude the optimization of \( p(\mathbf{y}_1 | \mathbf{x}, \mathbf{z}, \mathbf{y}_0, \mathbf{f}) \) and use tree-based generation after each RL iteration (\( N \)=16), focusing on instruction following with the common preference in Table~\ref{tab:preference}.}
    \label{fig:ablation_improve}
\end{figure*}
In our ablation study, we evaluate the impact of removing the optimization of \( p(\bm{y}_1 | \bm{x}, \bm{z}, \bm{y}_0, \bm{f}) \) during \emph{PRS} RL training. We apply tree-based generation after each RL iteration to sample outputs. As shown in Fig.~\ref{fig:ablation_improve}, skipping this optimization results in lower rewards. Additionally, the performance declines in the ablated baseline starting from the third RL iteration, highlighting the importance of this component.

\section{Details of Preference Adaptation}\label{sec:preference_adaptation}
We assess the adaptability of models to input preferences by specifying a desired output preference after the input question. 
Our comparison focuses on the methods of \emph{PRS}, PRand, and Rand, examining how different sampling approaches influence the alignment of models with user preferences. 
Both \emph{PRS} and PRand are designed to train models explicitly to adhere to given preferences, unlike Rand which does not incorporate preference training. 

Given the impracticality of evaluating every conceivable preference, we define five types of preferences which are \emph{conciseness}, \emph{thoroughness}, \emph{clarity}, \emph{professional tone}, and \emph{humorous tone}. 
To ensure robust generalization, we use GPT-4 to formulate 20 unique expressions for each preference category. Each preference within a category conveys the same underlying preference but is expressed in different styles. We present one expression for each category as follows:
\begin{itemize}[leftmargin=0.4cm, itemsep=-3pt, topsep=2pt]
    \item \textbf{Conciseness}: I prefer short and sweet responses.
    \item \textbf{Thoroughness}: I prefer responses that delve deeply into the topic and provide comprehensive
insights.
    \item \textbf{Clarity}: I prefer responses that are clear and straightforward, avoiding any unnecessary complexity or jargon.
    \item \textbf{Professional Tone}: I prefer responses that maintain a professional tone throughout the conversation. 
    \item \textbf{Humorous Tone}: I prefer responses with a side of humor, please. 
\end{itemize}
Sample preferences are shown in Table~\ref{tab:example_preference_diff_category}. 

We sample 100 test cases from AlpacaEval for evaluation. We conduct a head-to-head comparison between the models tuned by \emph{PRS}, \emph{PRand}, and Rand. For each category, we randomly sample an expression from 20 candidates and append the preference after the input prompt when generating a response. We also use GPT-4 as the evaluator, adapting the evaluation code from AlpacaEval. We specifically indicate a preference type for GPT-4 to choose a better-aligned response from two candidates.

\section{Prompts used in \emph{PRS}}\label{sec:prompt_pres}
\begin{figure*}[t]
\centering
\includegraphics[width=0.6\textwidth]{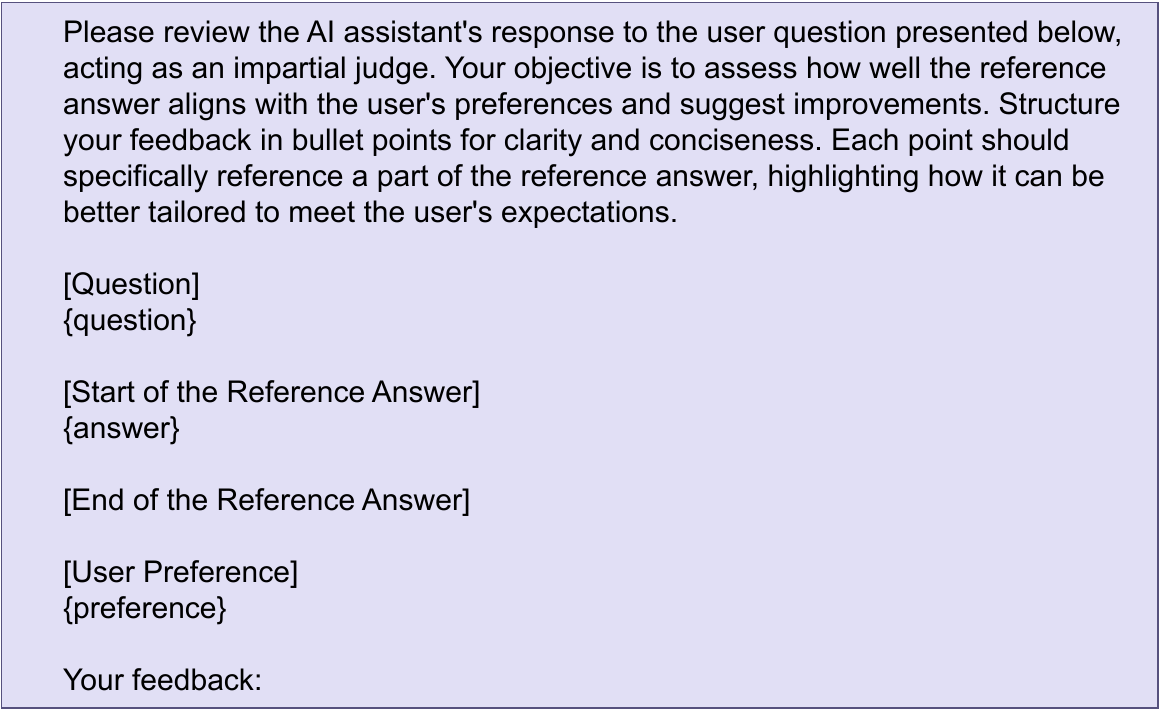} 
\caption{Prompt template for feedback generation, for the task of instruction following.}
\label{fig:prompt_feedback_gen}
\end{figure*}

\begin{figure*}[t]
\centering
\includegraphics[width=0.6\textwidth]{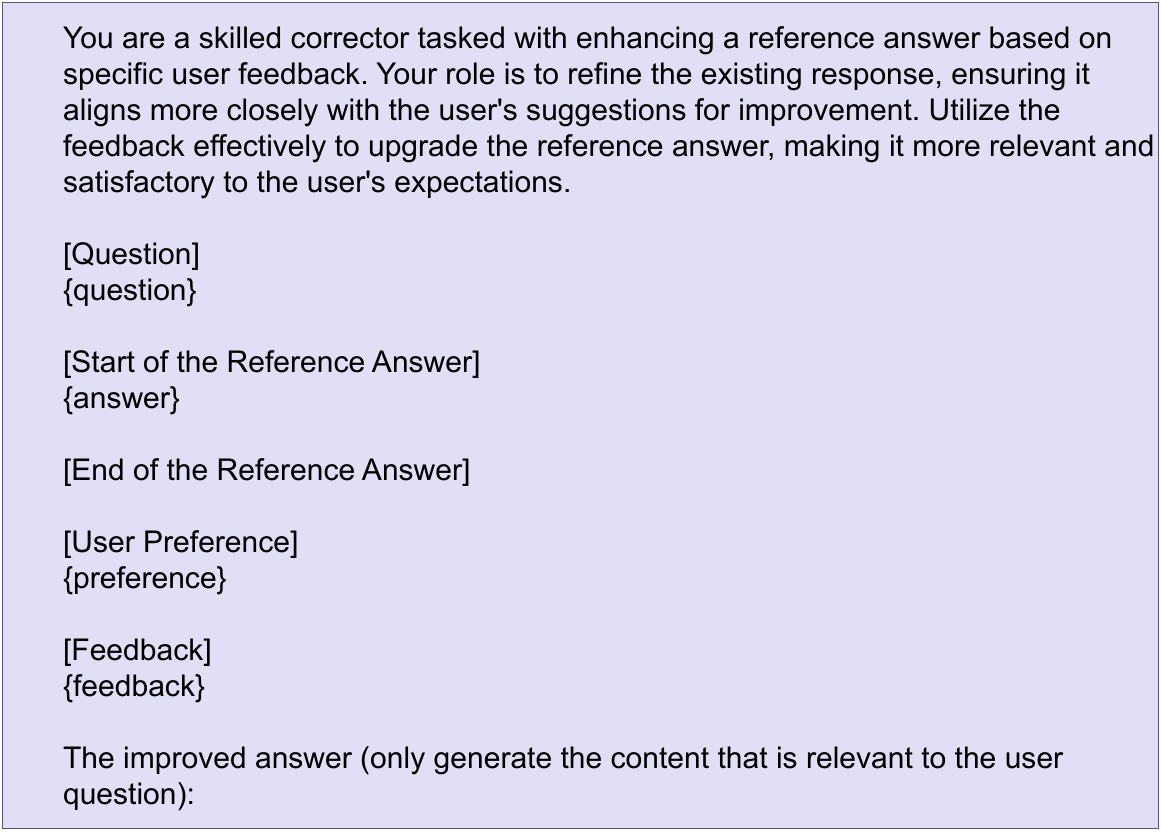} 
\caption{Prompt template for refinement with feedback, for the task of instruction following.}
\label{fig:prompt_refine_w_feedback}
\end{figure*}

\begin{figure*}[t]
\centering
\includegraphics[width=0.6\textwidth]{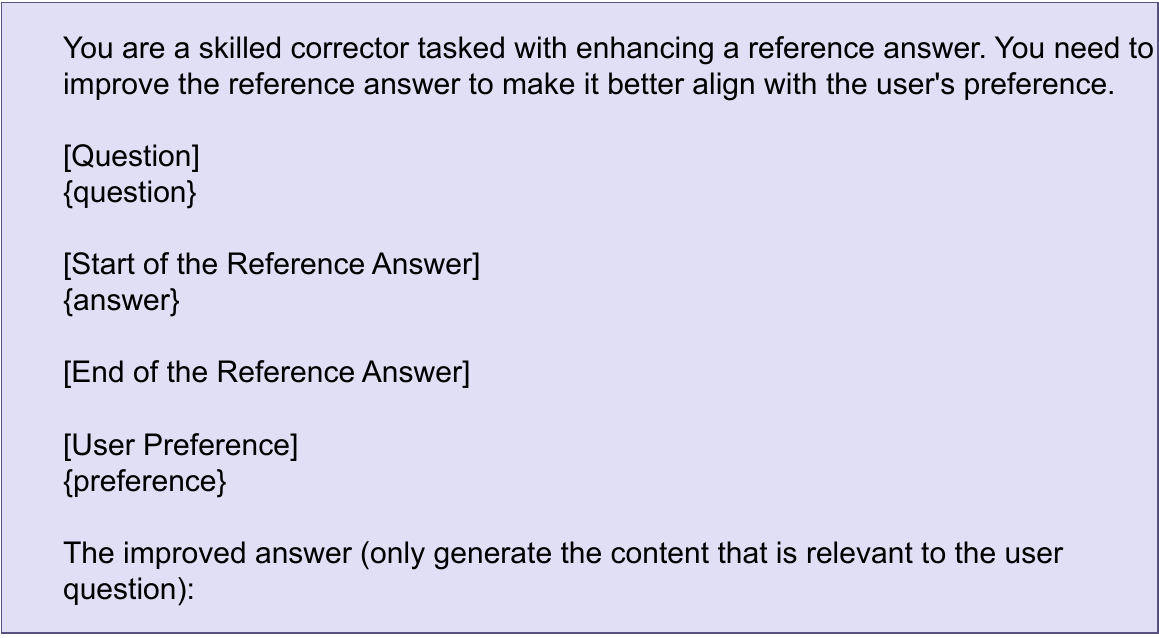} 
\caption{Prompt template for direct refinement without feedback, for the task of instruction following.}
\label{fig:prompt_refine_wo_feedback}
\end{figure*}

\begin{figure*}[t]
\centering
\includegraphics[width=0.6\textwidth]{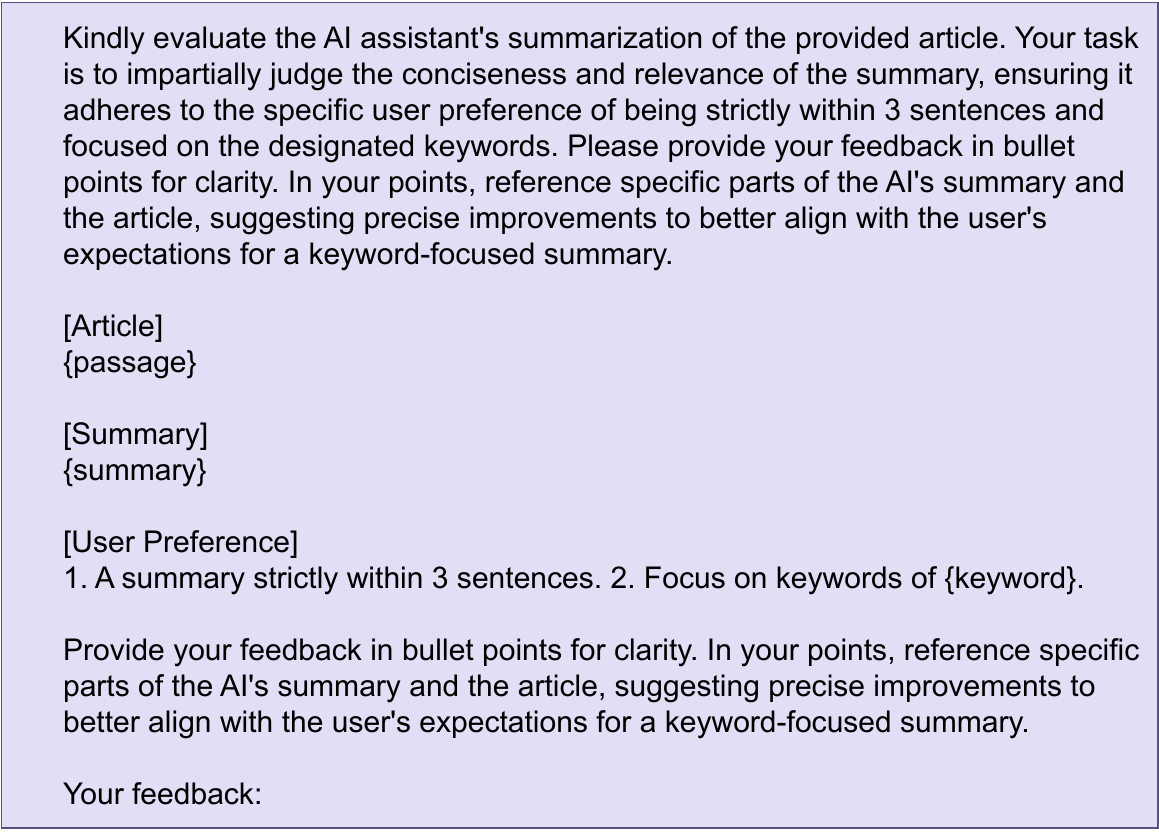} 
\caption{Prompt template for feedback generation, for the task of summarization.}
\label{fig:prompt_feedback_gen_summarization}
\end{figure*}

\begin{figure*}[t]
\centering
\includegraphics[width=0.6\textwidth]{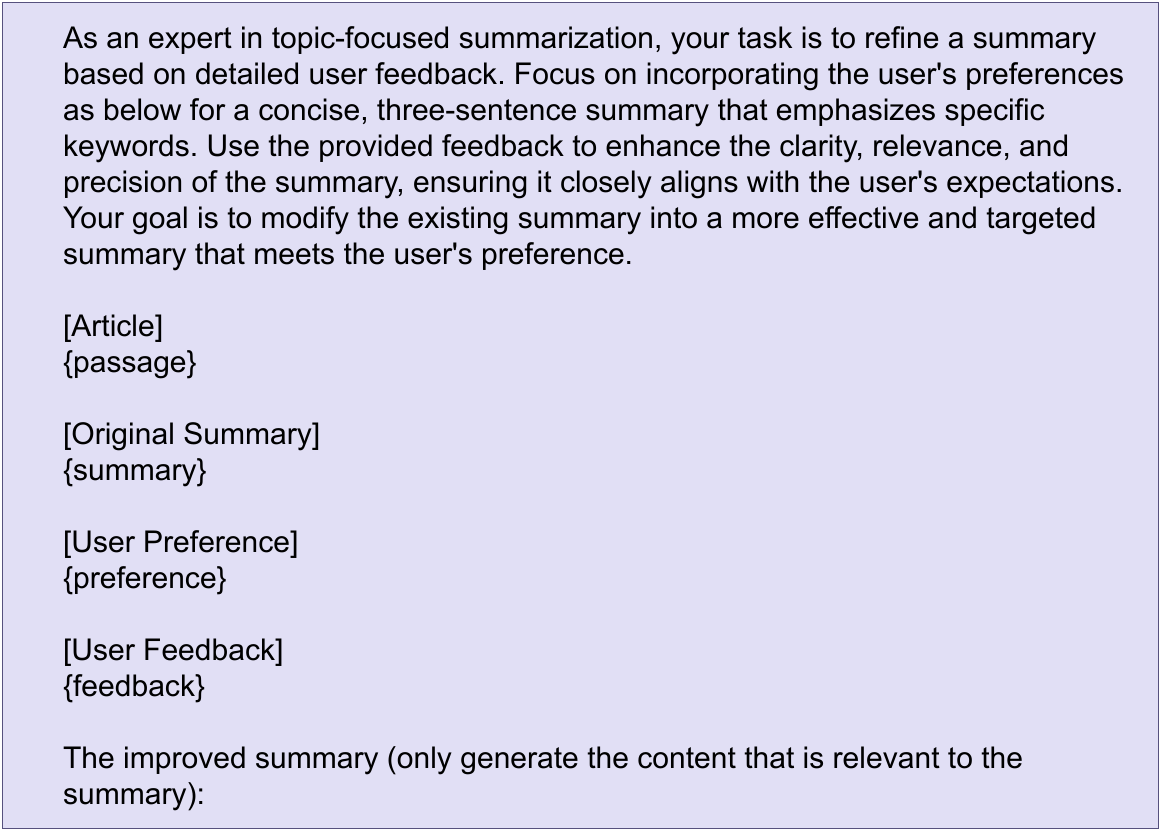} 
\caption{Prompt template for refinement with feedback, for the task of summarization.}
\label{fig:prompt_refine_w_feedback_summarization}
\end{figure*}

For instruction following, we present the prompt used for generating feedback in Fig.~\ref{fig:prompt_feedback_gen}, the prompt for revising the response with feedback in Fig.~\ref{fig:prompt_refine_w_feedback}, and the prompt for direct refinement without feedback in Fig.~\ref{fig:prompt_refine_wo_feedback}. For keyword-focused document summarization, the prompt for providing feedback is shown in Fig.~\ref{fig:prompt_feedback_gen_summarization} and the prompt for refinement with feedback is shown in Fig.~\ref{fig:prompt_refine_w_feedback_summarization}. 

\section{Details of Preference Generation}\label{appendix:preference_gen}
\begin{figure*}[t]
\centering
\includegraphics[width=0.45\textwidth]{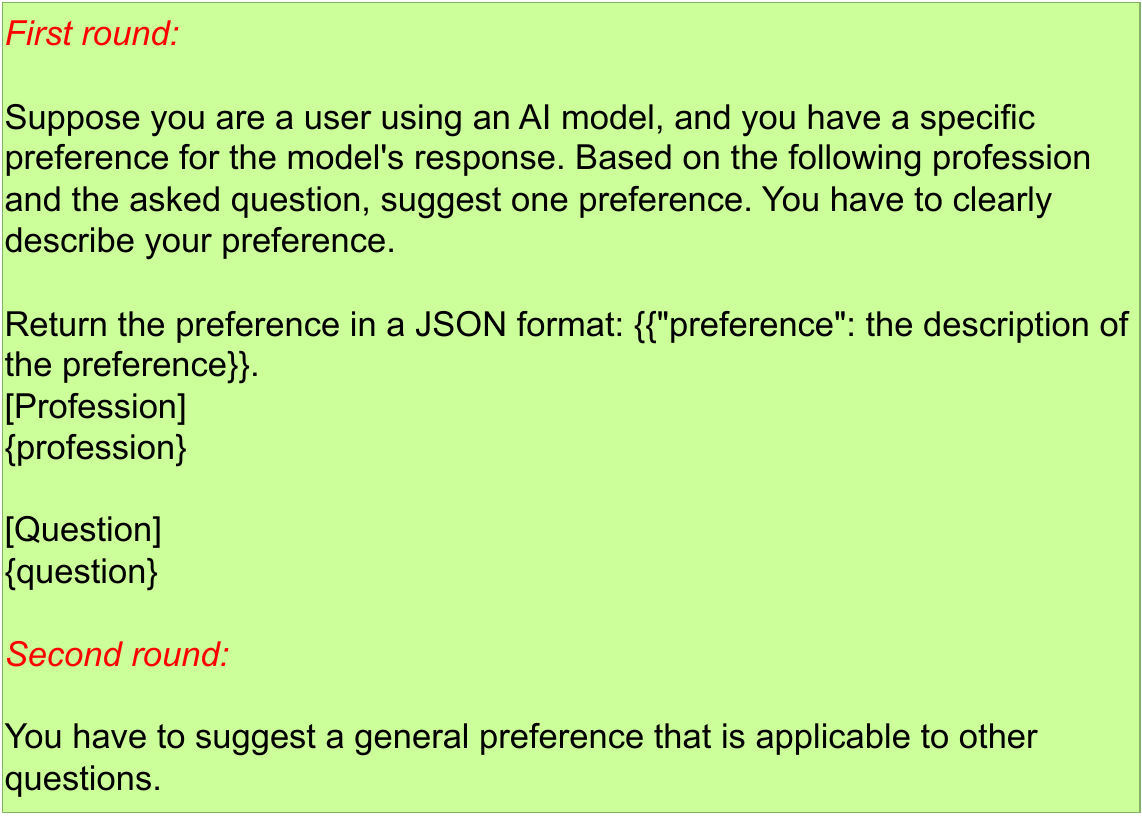} 
\caption{Prompt template for generating preferences, for the task of instruction following.}
\label{fig:prompt_preference_gen}
\end{figure*}

\noindent{\textbf{List of Professions.}} \ We have 222 professions in total divided into different categories:
\begin{itemize}[leftmargin=0.4cm, itemsep=-3pt, topsep=2pt]
    \item \textbf{Information Technology:} Software Developer, System Administrator, Network Engineer, Database Administrator, IT Support Specialist, Security Analyst, Data Scientist, Business Analyst, IT Project Manager, Web Developer, Quality Assurance Analyst, DevOps Engineer, Cloud Solutions Architect, IT Consultant, UX/UI Designer, Technical Writer, Mobile App Developer, Data Analyst, Information Systems Manager, and Computer Systems Analyst.

    \item \textbf{Business:} Accountant, Financial Analyst, Marketing Manager, Sales Representative, Business Analyst, Operations Manager, Human Resources Specialist, Management Consultant, Entrepreneur, Product Manager, Project Manager, Supply Chain Analyst, Customer Service Representative, Business Development Manager, and Data Analyst.

    \item \textbf{Retail:} Cashier, Sales Associate, Store Manager, Assistant Store Manager, Retail Merchandiser, Customer Service Representative, Stock Clerk, Visual Merchandiser, Loss Prevention Officer, Department Manager, Buyer, Inventory Control Specialist, Store Owner, E-commerce Specialist, and Retail Sales Consultant.

    \item \textbf{Health and Social Work:} Doctor, Nurse, Social Worker, Physical Therapist, Occupational Therapist, Dentist, Pharmacist, Clinical Psychologist, Counselor, Healthcare Administrator, Medical Laboratory Technician, Home Health Aide, Radiologic Technologist, Dietitian, Speech-Language Pathologist, Medical Assistant, Public Health Specialist, Chiropractor, Optometrist, Mental Health Technician, and Health Educator.

    \item \textbf{Transportation:} Truck Driver, Delivery Driver, Bus Driver, Taxi Driver, Pilot, Flight Attendant, Railway Conductor, Train Operator, Ship Captain, Sailor, Air Traffic Controller, Logistics Coordinator, Supply Chain Manager, Freight Agent, Transportation Planner, Transportation Engineer, Bicycle Courier, Warehouse Worker, Forklift Operator, and Aircraft Maintenance Technician.

    \item \textbf{Writing and Creative Arts:} Author, Screenwriter, Journalist, Editor, Copywriter, Content Creator, Blogger, Playwright, Poet, Graphic Designer, Illustrator, Animator, Photographer, Videographer, Filmmaker, Actor, Director, Producer, Musician, Composer, Visual Artist, Sculptor, Painter, Dancer, Choreographer, and Performance Artist.

    \item \textbf{Broadcasting and Entertainment:} Actor, Director, Producer, Screenwriter, Cinematographer, Film Editor, Broadcast Journalist, Television Presenter, Radio Presenter, News Anchor, Camera Operator, Sound Engineer, Lighting Technician, Production Designer, Makeup Artist, Costume Designer, Animator, Visual Effects Artist, Music Composer, Singer, Musician, Stand-up Comedian, Talent Manager, Casting Director, and Stage Manager.

    \item \textbf{Law and Order:} Lawyer, Paralegal, Judge, Police Officer, Correctional Officer, Detective, Prosecutor, Public Defender, Legal Assistant, Bailiff, Criminologist, Forensic Scientist, Court Reporter, Private Investigator, Legal Secretary, Probation Officer, Court Clerk, Security Guard, Prison Warden, and Compliance Officer.

    \item \textbf{Sports and Recreation:} Athlete, Coach, Sports Agent, Physical Therapist, Personal Trainer, Referee/Umpire, Sports Journalist, Sportscaster, Fitness Instructor, Recreation Worker, Athletic Trainer, Sports Photographer, Sports Marketing Specialist, Sports Psychologist, Sports Nutritionist, Gym Manager, Outdoor Activity Coordinator, Sports Statistician, Team Manager, and Scout.

    \item \textbf{Education:} Teacher, School Principal, School Counselor, Librarian, Teaching Assistant, Education Administrator, Instructional Coordinator, Special Education Teacher, University Professor, Tutor, Educational Consultant, College Admissions Officer, Academic Advisor, School Psychologist, Education Policy Analyst, Curriculum Developer, Education Researcher, Literacy Coach, Physical Education Teacher, and ESL Teacher.

    \item \textbf{Scientific Research:} Research Scientist, Laboratory Technician, Research Assistant, Data Analyst, Statistician, Biologist, Chemist, Physicist, Biochemist, Clinical Research Associate, Epidemiologist, Environmental Scientist, Geneticist, Microbiologist, Astrophysicist, Geologist, Postdoctoral Researcher, Principal Investigator, Research Fellow, and Scientific Writer.

\end{itemize}

\noindent{\textbf{Preference Annotation.}} \ 
\begin{figure*}[t]
\centering
\includegraphics[width=0.7\textwidth]{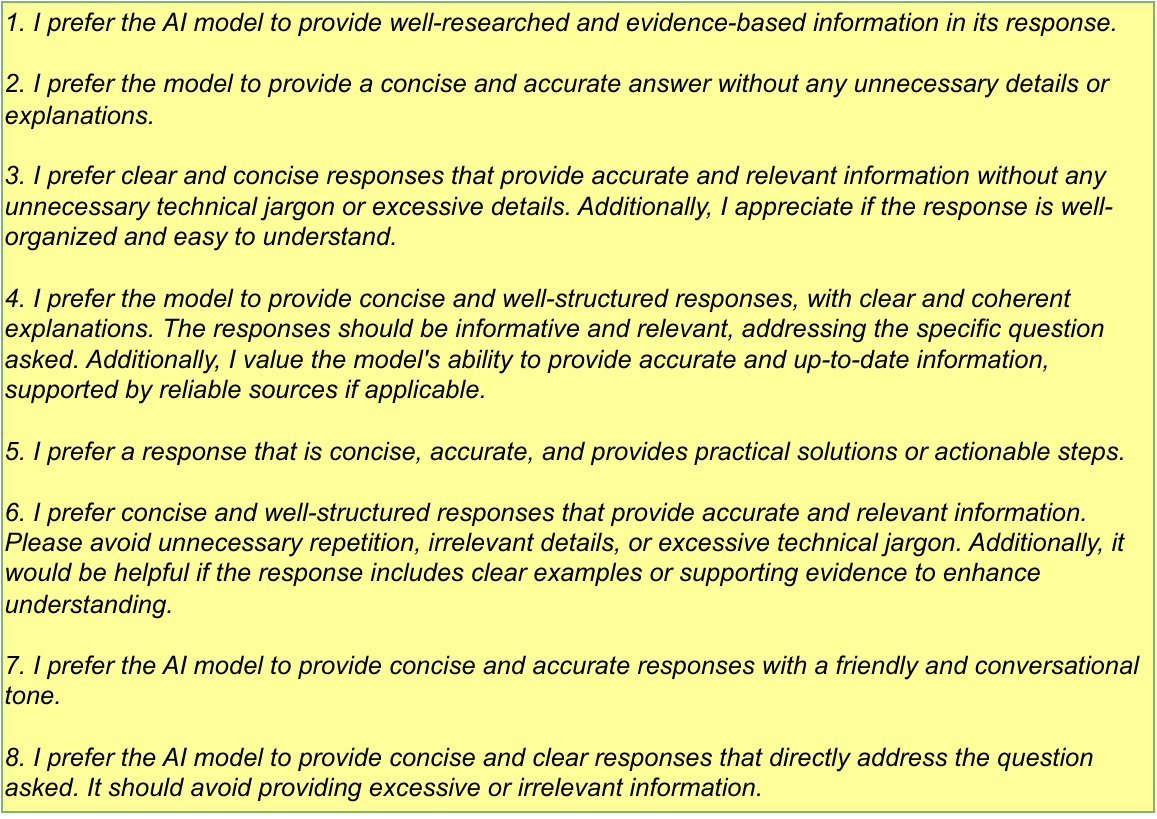} 
\caption{Sample annotated preferences by GPT-3.5-turbo.}
\label{fig:example_preference}
\end{figure*}
We use GPT-3.5-turbo to generate the preferences. For each prompt from Alpaca-GPT4, we use the template in Fig.~\ref{fig:prompt_preference_gen} to generate the preference, where the generation is conditioned on the question and a profession name. The profession name is randomly selected from the profession name list. After obtaining a preference, we further prompt GPT-3.5-turbo to revise its output to make the generated preference general and applicable to different questions. In Fig.~\ref{fig:example_preference}, we present a variety of generated preferences, illustrating the diversity in the preferences that the method can produce.

\begin{table*}[t]
\centering
\resizebox{0.7\textwidth}{!}{%
\begin{tabular}{lp{12cm}}
\toprule
  & \textbf{Conciseness}                                    \\ \midrule
1 & I prefer short and sweet responses.                     \\
2 & I prefer answers that are to the point.                 \\
3 & I prefer concise explanations, no fluff.                \\ \midrule
  & \textbf{Thoroughness}                               \\ \midrule
1  & I prefer responses that delve deeply into the topic and provide comprehensive insights                        \\
2  & I prefer when the information is thorough and covers all aspects, leaving no stone unturned.                  \\
3  & I prefer a detailed exposition, with rich context and nuanced explanations.                                   \\ \midrule
  & \textbf{Clarity}                                        \\ \midrule
1 & I prefer responses that are clear and straightforward, avoiding any unnecessary complexity or jargon.         \\
2 & I prefer that you explain things simply, as if you were talking to someone who's completely new to the topic. \\
3 & I prefer answers that are easy to understand and follow, without any convoluted explanations.                 \\ \midrule
  & \textbf{Professional Tone}                              \\ \midrule
1  & I prefer responses that maintain a professional tone throughout the conversation.                             \\
2  & I prefer that the language used is formal and professional in nature.                                         \\
3  & I prefer the communication to be strictly professional. \\ \midrule
  & \textbf{Humorous Tone}                                  \\ \midrule
1 & I prefer responses with a side of humor, please.        \\
2 & I prefer my information served with a chuckle.          \\
3 & I prefer answers that come with a comedic twist.        \\ \bottomrule
\end{tabular}%
}
\caption{Sample preferences with different expressions for each category. Three examples are shown in each category.}
\label{tab:example_preference_diff_category}
\end{table*}

\begin{table*}[t]
\centering
\resizebox{\textwidth}{!}{%
\begin{tabular}{|l|p{16cm}|}
\hline
 &
  Do dinosaurs really exist? \\ \hline
Rand &
  Yes, dinosaurs are a group of extinct reptiles that lived on Earth during the Mesozoic Era, which lasted from about 250 million to 66 million years ago. The term "dinosaur" was coined by Sir Richard Owen in 1842, and it means "terrible lizard." 
  
  There is overwhelming evidence that dinosaurs existed, including fossils found all over the world, as well as scientific research and analysis of these fossils. Paleontologists have discovered thousands of different species of dinosaurs, ranging in size from tiny, bird-like creatures to massive giants like Tyrannosaurus rex. 
  
  While there is no doubt that dinosaurs existed, there is still much that we don't know about them, such as how they behaved, what their environments were like, and why they went extinct. However, through continued research and discovery, scientists are constantly learning more about these fascinating creatures. \\ \hline
 &
  Do dinosaurs really exist? 
  
  I prefer responses that are informative, precise, creative, detailed, relevant, and in-depth. \\ \hline
Rand + p &
  Yes, dinosaurs existed. They were a group of animals that lived on Earth during the Mesozoic era, which lasted from about 250 million to 66 million years ago. The first known dinosaurs appeared around 230 million years ago and went extinct at the end of the Cretaceous period, about 66 million years ago.
  
  There are many different types of dinosaurs, including Tyrannosaurus rex, Triceratops, Velociraptor, Stegosaurus, Diplodocus, and Brachiosaurus. These animals ranged in size from small, bipedal carnivores like Velociraptor to massive, quadrupedal herbivores like Apatosaurus. Some dinosaurs had feathers, while others had armor or spikes on their bodies.
  
  Dinosaur fossils have been found all over the world, including North America, South America, Africa, Europe, Asia, and Australia. These fossils provide evidence of what these animals looked like and how they lived. Scientists use this information to learn about the evolution of life on Earth and to understand the history of our planet. \\ \hline
PRand &
  Yes, the dinosaurs really existed. They were a diverse group of reptiles that lived on Earth during the Mesozoic Era, which lasted from about 251 to 66 million years ago. The first dinosaurs appeared in the Triassic period and went extinct at the end of the Cretaceous period.
  
  Dinosaurs were large, land-dwelling animals that varied greatly in size, shape, and behavior. Some were herbivores (plant-eaters) like the long-necked sauropods such as Brachiosaurus and Diplodocus, while others were carnivores (meat-eaters) like Tyrannosaurus rex and Velociraptor. Some dinosaurs could fly, like Pteranodon, while others could swim, like Spinosaurus.
  
  There have been many discoveries of fossilized remains of dinosaurs around the world, providing evidence of their existence. These include complete skeletons, tracks, eggs, nests, and even preserved soft tissues. Scientists use these fossils to study the evolution, anatomy, and behavior of dinosaurs, helping us understand more about life on Earth millions of years ago. \\ \hline
\emph{PRS} &
  Yes, dinosaurs are a group of extinct reptiles that lived on Earth during the Mesozoic Era, which lasted from approximately 252 to 66 million years ago. They first appeared in the fossil record during the Triassic Period and became extinct at the end of the Cretaceous Period, along with many other species due to a massive asteroid impact.
  
  Dinosaurs were diverse and came in various shapes and sizes, ranging from small bird-like creatures to gigantic herbivores and carnivores. Some well-known dinosaurs include Tyrannosaurus rex, Triceratops, Stegosaurus, and Velociraptor. These animals had unique features such as horns, spikes, plates, and feathers that helped them adapt to their environment.
  
  The discovery of dinosaur fossils has played a crucial role in our understanding of the history of life on Earth. Paleontologists have found thousands of dinosaur fossils all over the world, providing valuable insights into how these creatures evolved over time and what kind of environments they inhabited. By studying dinosaur fossils, we can learn about the geological changes that occurred throughout the Mesozoic Era, including shifts in climate, sea level, and landmass distribution.
  
  Overall, there is overwhelming evidence supporting the existence of dinosaurs. Their remains have been found in rocks dated back millions of years ago, and their fossils show distinctive characteristics that differentiate them from other animals alive today. Therefore, it is reasonable to conclude that dinosaurs were real animals that once roamed our planet. \\ \hline
\end{tabular}%
}
\caption{Example 1 demonstrates outputs using different sampling methods to align the language model.}
\label{tab:example-output-1}
\end{table*}

\begin{table*}[t]
\centering
\resizebox{\textwidth}{!}{%
\begin{tabular}{|l|p{16cm}|}
\hline
 &
  Decide whether children can watch the given movie. Mention what makes it inappropriate if you think they can't. 
  
  The Dark Knight (2008) \\ \hline
Rand &
  The Dark Knight is a superhero film that follows the story of Batman as he battles against the villainous Joker. While it does not contain any explicit content, the movie has a dark and intense tone that may be too mature for younger children. Additionally, some scenes feature violence and action sequences that could be frightening for young viewers. Overall, I would recommend this movie for teenagers or adults who are comfortable with its themes and imagery. \\ \hline
 &
  Decide whether children can watch the given movie. Mention what makes it inappropriate if you think they can't.
  
  The Dark Knight (2008)  
  
  I prefer responses that are informative, precise, creative, detailed, relevant, and in-depth. \\ \hline
Rand + p &
  The Dark Knight is a superhero film that follows the story of Batman as he tries to take down the criminal mastermind, the Joker. The movie has received critical acclaim for its performances, direction, and themes, but it also contains some mature content that may not be suitable for younger audiences.
  
  While children can certainly watch The Dark Knight, parents should exercise caution and consider their child's age before allowing them to do so. The movie features intense violence, language, and themes of death and tragedy that may be too mature for younger viewers. Additionally, the Joker's character is particularly disturbing and may frighten young children.
  
  Overall, while The Dark Knight is an excellent movie, parents should use their discretion when deciding whether or not to allow their children to watch it. \\ \hline
PRand &
  The Dark Knight is a superhero film based on the DC Comics character Batman. The movie follows Bruce Wayne, who assumes the identity of Batman to protect Gotham City from criminals. The film explores themes such as crime, corruption, and morality. It was rated PG-13 for intense sequences of violence and menace, some sexual content, and language.
  
  In my opinion, children can watch The Dark Knight with parental guidance. While the movie has some violent scenes, they are not overly graphic or gory. There is also some sexual content, but it is not explicit or gratuitous. The language used in the movie is also not too crude or offensive. However, parents should be aware that the movie deals with mature themes and may be too complex for younger audiences to fully understand. Additionally, the film's runtime is over two hours long, which could make it difficult for younger children to sit through without getting restless. Overall, I believe that older children (age 13 and above) can enjoy The Dark Knight with proper supervision and discussion with their parents about its themes and content. \\ \hline
\textit{PRS} &
  The Dark Knight is a 2008 superhero movie based on the DC Comics character Batman. The film features intense violence, profanity, and mature themes that may not be appropriate for children.
  
  Violence: The Dark Knight contains numerous scenes of violent crime, including gunfire, explosions, and hand-to-hand combat. These scenes can be particularly disturbing for younger viewers who may not understand the context or consequences of such violence.
  
  Profanity: The film also includes frequent use of strong language, including profanities like \textbackslash{}"f**k\textbackslash{}" and \textbackslash{}"sh*t.\textbackslash{}" This type of language can be jarring for some parents who want to shield their children from hearing such words.
  
  Mature Themes: In addition to its violence and language, The Dark Knight deals with complex issues like terrorism, corruption, and moral ambiguity. These themes can be challenging for young children to process and may require additional discussion or explanation from parents after watching the movie.
  
  Overall, while The Dark Knight is an excellent film for older teenagers and adults, it may not be suitable for younger children due to its intense content and mature themes. Parents should exercise caution when deciding whether to allow their children to watch this movie. \\ \hline
\end{tabular}%
}
\caption{Example 2 demonstrates outputs using different sampling methods to align the language model.}
\label{tab:example-output-2}
\end{table*}

\section{Sample Outputs of Different Baselines}
We display sample outputs in Tables~\ref{tab:example-output-1} and \ref{tab:example-output-2}.

\end{document}